\documentclass[letterpaper]{article} 
\usepackage{aaai23}  
\usepackage{times}  
\usepackage{helvet}  
\usepackage{courier}  
\usepackage[hyphens]{url}  
\usepackage{graphicx} 
\usepackage{natbib}  
\usepackage{caption} 
\usepackage{algorithm}
\usepackage{algorithmic}

\usepackage{newfloat}
\usepackage{listings}
\DeclareCaptionStyle{ruled}{labelfont=normalfont,labelsep=colon,strut=off} 
\floatstyle{ruled}
\newfloat{listing}{tb}{lst}{}
\floatname{listing}{Listing}
\usepackage{booktabs}
\usepackage{amsmath,amssymb} 
\newcommand{\bsb}[1]{\boldsymbol{#1}}

\title{ContraFeat: Contrasting Deep Features for Semantic Discovery}
\author{
    Xinqi Zhu, Chang Xu, Dacheng Tao
}
\affiliations{
    The University of Sydney, Australia\\
    xzhu7491@uni.sydney.edu.au, c.xu@sydney.edu.au, dacheng.tao@gmail.com

}

\usepackage{bibentry}

\begin{document}

\maketitle

\begin{abstract}
    StyleGAN has shown strong potential
    for disentangled semantic control, thanks to its special design
    of multi-layer intermediate latent variables.
    However, existing semantic discovery methods on StyleGAN
    rely on manual selection of modified latent layers
    to obtain satisfactory manipulation results,
    which is tedious and demanding.
    In this paper, we propose a model that automates this process
    and achieves state-of-the-art semantic discovery performance.
    The model consists of an attention-equipped navigator module
    and losses contrasting deep-feature changes.
    We propose two model variants, with one contrasting
    samples in a binary manner,
    and another one contrasting
    samples with learned prototype variation patterns.
    The proposed losses are defined with pretrained deep features,
    based on our assumption that the
    features can implicitly reveal the desired semantic
    structure including consistency and orthogonality.
    Additionally, we design two metrics to quantitatively evaluate
    the performance of semantic discovery methods on FFHQ dataset,
    and also show that disentangled representations can be
    derived via a simple training process.
    Experimentally, our models can obtain state-of-the-art semantic
    discovery results without relying on latent layer-wise manual selection,
    and these discovered semantics can be used to manipulate real-world images.
\end{abstract}

\section{Introduction}
\label{sec:introduction}

\begin{figure}[t]
    \begin{center}
       \includegraphics[width=\linewidth]{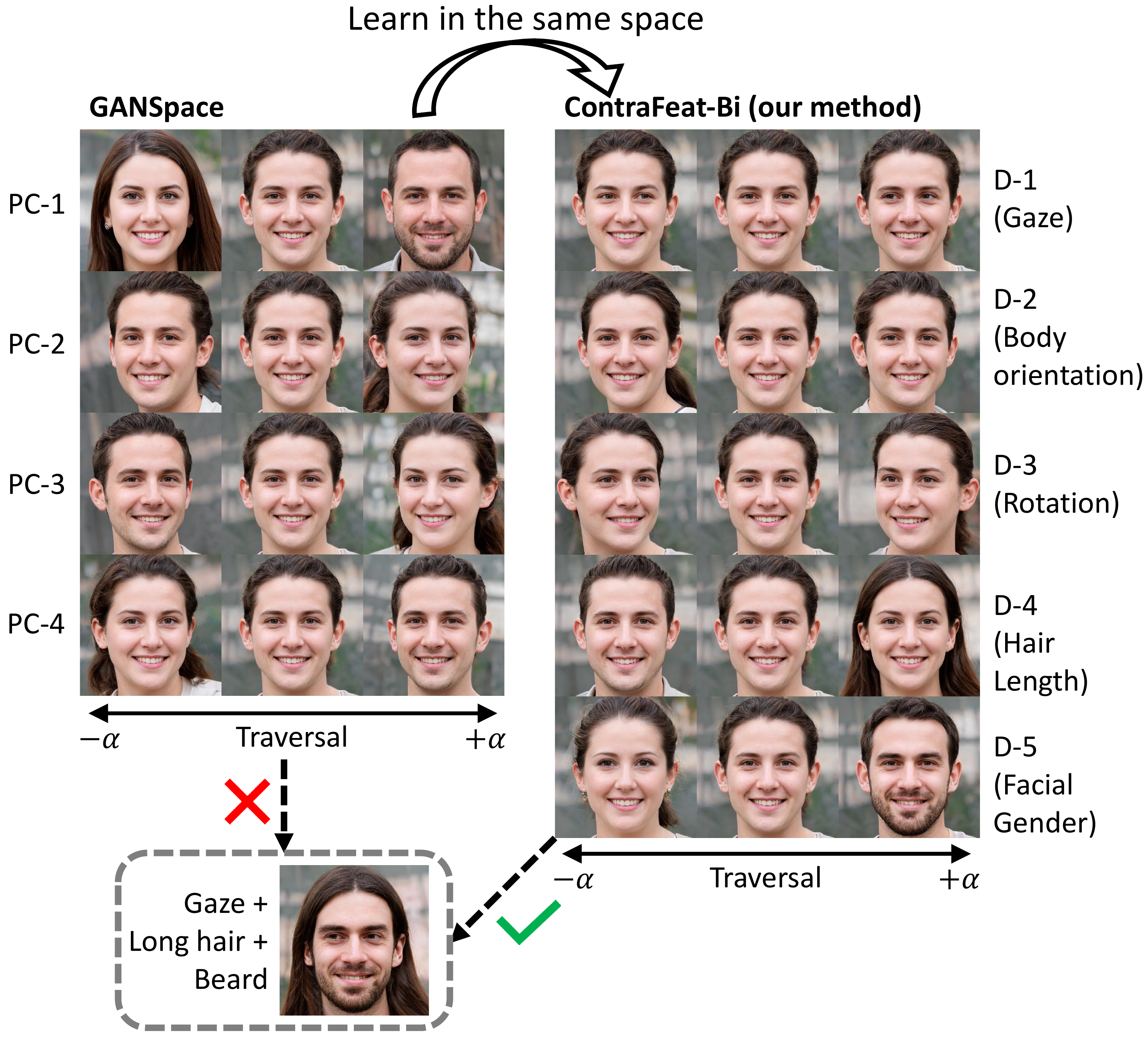}
    \end{center}
        \caption{Discovered semantics using Ganspace
        \cite{hrknen2020ganspace}
        and our ContraFeat model in the same $\mathcal{W}$ subspace (the subspace
        defined by the top 4 principal components found by Ganspace) of a
        StyleGAN2 network. No manual $\mathcal{W}$-space layer
        selection is conducted.
        In this example, our ContraFeat model
        discovers more disentangled semantics than Ganspace,
        including the decomposed gender and hair length attributes
        and the less obvious gaze change.
        This enables the construction of an out-of-distribution
        sample \emph{a man with long hair and beard gazing at the left}
        shown in the bottom-left, which is not possible using Ganspace.}
    \label{fig:teaser}
\end{figure}
Deep generative models such as GANs \cite{Goodfellow2014GenerativeAN}
have shown superior ability in generating
photo-realistic images \cite{Karras2017ProgressiveGO,Karras2020AnalyzingAI,Brock2018LargeSG}.
Besides improving fidelity in generation,
many works seek to augment these models with
interpretability \cite{Yan2016Attribute2ImageCI,Chen2016InfoGANIR,
Higgins2017betaVAELB,bau2019gandissect,Shen2019InterpretingTL},
which fertilizes
applications such as controllable generation
\cite{Kulkarni2015DeepCI,Lample2017FaderNM,Lee2018DiverseIT,
Xing2019UnsupervisedDO},
domain adaptation \cite{Peng2019DomainAL,Cao2018DiDADS}, machine learning
fairness \cite{Creager2019FlexiblyFR,Locatello2019OnTF}, etc.
Yet this goal can be typically achieved via
supervised learning \cite{Kingma2014SemisupervisedLW,
Dosovitskiy2014LearningTG,Kulkarni2015DeepCI,
Lample2017FaderNM,Shen2019InterpretingTL},
another recent popular branch of work
approaches this target by \emph{semantic discovery}.
Instead of training a fresh generator with labels, these semantic discovery
methods \cite{voynov2020unsupervised,hrknen2020ganspace,wang2021aGANGeom,
shen2021closedform,
yuksel2021latentclr} seek semantically meaningful directions in the
latent space of pretrained GANs without supervision.
Among popular GANs, StyleGAN has shown great
potential for disentangling semantics, thanks to its intermediate
$\mathcal{W}$ space design with a multi-layer structure.
This paper focuses on StyleGAN to develop a
tailored semantic discovery model in its $\mathcal{W}$ space.

Although many existing models \cite{hrknen2020ganspace,shen2021closedform,
yuksel2021latentclr} have demonstrated ability in discovering
semantically meaningful variations in StyleGANs, these methods
share a common problem: they
rely on manual layer-wise selection in the $\mathcal{W}$ space.
As mentioned in \cite{hrknen2020ganspace}, this layer-wise selection
is important for satisfactory manipulation results since entanglement can frequently happen 
in the discovered directions without this process.
Unfortunately, this manual selection process is tedious and
demanding because it requires humans
to inspect and evaluate a large number of layer-wise
modification combinations in the $\mathcal{W}$ latent space.

There are two major reasons causing this problem
in existing models:
the first is the lack of a module designed for this layer-selection process;
the second is the lack of a criterion that can substitute human
intelligence to guide this selection.
To solve the first problem, we develop a module to predict
layer-wise attentions in the $\mathcal{W}$ space which simulates
the selection of latent layers to apply discovered directions.
For the second problem, we adopt a pretrained neural network to
serve as the selective guidance.
This is inspired by existing observation that the perceptual
judgment of deep network features
largely agrees with human judgment \cite{zhang2018perceptual}.
In this paper, we make the hypothesis that this
perceptual agreement also extends to semantic structure,
e.g. similarity in \emph{semantic changes} judged by humans is agreed
by neural networks, and difference as well.
Technically, we develop a contrastive approach to encourage
the discovery of semantics that will cause perceptually
consistent and orthogonal changes in the pretrained feature maps.
We propose two variants of our model, with one contrasting
samples in a binary manner, and another one contrasting
samples with learned prototype variation patterns.
With the proposed models,
we can discover more disentangled semantics on StyleGANs than
existing methods without relying
on manual layer-wise selection on $\mathcal{W}$ space.
As shown in Fig. \ref{fig:teaser}, both methods (Ganspace
\cite{hrknen2020ganspace} and our model)
are applied to the same $\mathcal{W}$ space, and ours can discover
cleaner semantics than the Ganspace model.
Note that our model decomposes the hair length and gender attributes into
two controllable factors, thus it can manipulate the latent code
to generate an out-of-distribution face sample with long-hair and beard
(shown in the bottom-left), which is not possible using Ganspace.

Our main contributions are summarized as follows:
\begin{itemize}
        \item We propose to substitute the manual layer-wise selection procedure
            in the $\mathcal{W}$ space of StyleGANs by a learnable attention module.
        \item We propose to use the deep features in
            pretrained networks to guide the discovery of
            semantics, verifying our hypothesis
            that deep features perceptually
            agree with human judgment on the semantic structure.
        \item We design one of our proposed model variants to
            learn prototype patterns for
            discovered semantics, avoiding the requirement
            of sampling multiple changing samples during contrastive learning.
        \item State-of-the-art semantic discovery results can 
            be achieved with our models, and disentangled
            representations can be derived.
\end{itemize}

\section{Related Work}
\label{sec:related_work}
\textbf{Semantic Discovery.}
Generative adversarial networks \cite{Goodfellow2014GenerativeAN}
have demonstrated state-of-the-art performance
\cite{Karras2017ProgressiveGO,Brock2018LargeSG,
Karras2020ASG,Karras2020AnalyzingAI} in generating photo-realistic images.
Recently, works revealed that the latent spaces of well-trained GANs
can be interpreted with meaningful semantic directions
\cite{bau2019gandissect,Shen2019InterpretingTL}.
This phenomenon motivated the investigation of semantic properties in the
latent spaces \cite{hrknen2020ganspace,wang2021aGANGeom},
and the discovery of semantic directions
\cite{Plumerault2020Controlling,gansteerability,
voynov2020unsupervised,spingarn2021gan,shen2021closedform,yuksel2021latentclr}.
In both \cite{hrknen2020ganspace} and \cite{wang2021aGANGeom},
the latent space of GANs has been shown highly anisotropic,
revealing most semantic variations are encoded in a low-dimension
latent subspace.
Methods in \cite{Plumerault2020Controlling,gansteerability} successfully learn
latent directions to represent predefined transformations on images.
Without relying on optimizations, \cite{hrknen2020ganspace,
spingarn2021gan,shen2021closedform} discover
semantic attributes by finding the latent directions
that cause the most significant changes.
Unfortunately, this assumption is too simple
and the methods rely on manual adjustment in the latent space
to obtain satisfactory results.
Based on the idea that different latent directions should correspond
to different changes, \cite{voynov2020unsupervised} and
\cite{yuksel2021latentclr} adopted classification loss and contrastive loss
to encourage the separation of different directions respectively.
However, the former requires the training of a deep classifier,
and the later is limited to using early features
in the generator due to its dense computation in the contrastive loss.

\begin{figure*}[t]
    \begin{center}
       \includegraphics[width=\linewidth]{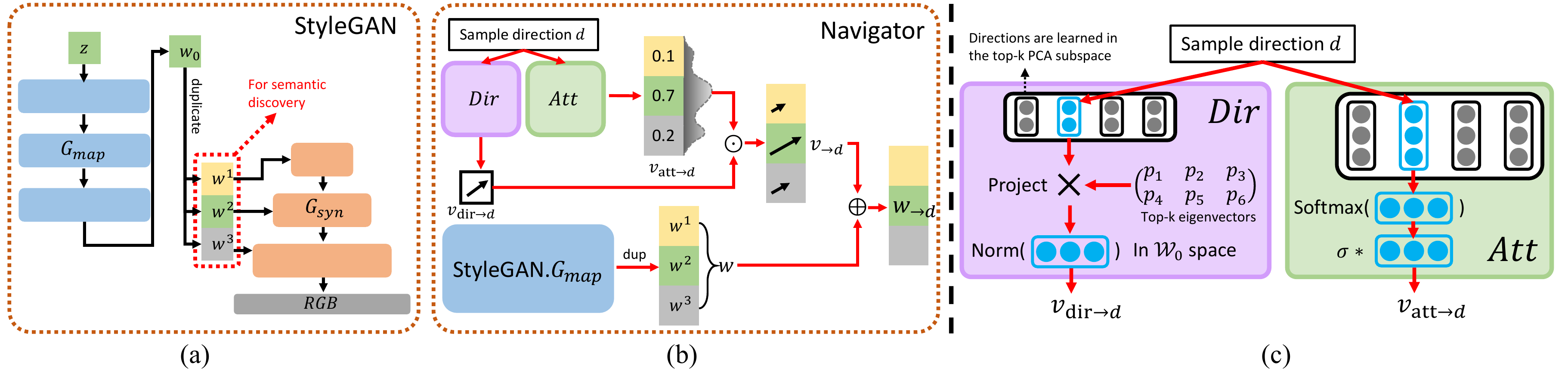}
    \end{center}
        \caption{(a)
        A simplified diagram of StyleGAN. The discovered
        latent directions are applied on the $\mathcal{W}$ space after duplication
        (red dotted box).
        (b) Overview of our
        proposed navigator. It consists of two branches.
        The direction branch ($Dir$) predicts
        directions $\bsb{v}_{\text{dir} \rightarrow d}$ in
        the $\mathcal{W}_0$ space and
        the attention branch ($Att$) predicts attentions on $\mathcal{W}$-space
        layers $\bsb{v}_{\text{att} \rightarrow d}$.
        The combined results $\bsb{v}_{\rightarrow d}$ are used to modify
        the latent codes $\bsb{w} \in \mathcal{W}$. 
        (c) Detailed illustration of
        $Dir$ and $Att$.}
    \label{fig:navigator}
\end{figure*}

\textbf{Disentanglement Learning.}
Another popular branch of work to unsupervisedly find
semantically meaningful changes is
by learning disentangled representations
\cite{Higgins2017betaVAELB,Chen2016InfoGANIR}.
In this domain, various regularization methods have been
proposed based on different assumptions, e.g. statistical independence
\cite{Burgess2018UnderstandingDI,
Kumar2017VariationalIO,Kim2018DisentanglingBF,chen2018isolating},
informativeness \cite{Chen2016InfoGANIR,Jeon2018IBGANDR,Xinqi_cvpr21},
separability \cite{Lin2019InfoGANCRDG},
and group decomposition \cite{Xinqi_liegroupvae_icml21}.
These models usually succeed at the cost of degraded generation quality
compared with the normal generative models.
In \cite{Locatello2018ChallengingCA}, Locatello et al. proved that
unsupervised disentanglement learning is impossible without additional
biases. In \cite{pmlr-v119-locatello20a}, Locatello et al. showed that
guaranteed disentanglement can be achieved with partial factor variation
information on some VAE variants \cite{ijcai2019-348,BouchacourtTN18}.
In this work, we show that state-of-the-art disentangled representations
can be derived using our semantic discovery model, and the biases
in our model include the perceptual contrastive property between semantics,
spatial separation, and their hierarchical abstractness difference.


\section{Method}
\label{sec:method}
In this section, we introduce our semantic discovery model
ContraFeat (\textbf{Contra}sting deep
\textbf{Feat}ures) in detail.

\subsection{Navigator}
\label{subsec:navigator}
In StyleGANs (a simplified diagram shown in Fig.
\ref{fig:navigator} (a)),
a normally sampled latent code $\bsb{z}$ is fed through a mapping network
$G_{map}$ to obtain an intermediate latent code $\bsb{w}_0$.
The $\bsb{w}_0 \in \mathcal{W}_0$ is then broadcast
as an extended code $\bsb{w} \in \mathcal{W}$ via duplication,
where $\mathcal{W}$ is the extended space $k$ times larger
than the original $\mathcal{W}_0$ space,
with $k$ determined by the depth of the synthesis network ($G_{syn}$).
We refer this duplication dimension as \emph{layer} of this $\mathcal{W}$ space.
Thanks to this hierarchical design, disentanglement
property emerges in StyleGANs \cite{Karras2020ASG,Karras2020AnalyzingAI}.
This extended latent space $\mathcal{W}$ plays an important role in
semantic editing, as works \cite{hrknen2020ganspace,shen2021closedform}
have shown that layer-wise editing in the $\mathcal{W}$ space achieves more
disentangled results. It is also discovered that real-world images
can be more accurately inverted into the extended $\mathcal{W}$ space
\cite{9008515,9157575} to perform semantic editing.

However, the existing works require manual selection of the modifiable
$\mathcal{W}$-space layers in order to curate satisfactory editing results.
To eliminate this tedious human effort, we propose a navigator network
to automate this procedure.
Specifically, the module conducts semantic direction prediction
(direction branch) and layer-wise
selection (attention branch) in the $\mathcal{W}$ space
(depicted in Fig. \ref{fig:navigator} (c) purple and green boxes respectively).
In our design, these two submodules are separated and thus it is possible
to replace the direction branch by other existing methods
to verify the compatibility of our layer-selection branch with other direction
prediction methods.

The direction branch
($Dir$ purple boxes in Fig. \ref{fig:navigator})
outputs $m$ latent directions
$\{\bsb{v}_{\text{dir} \rightarrow d}, d=1..m\}$
in the $\mathcal{W}_0$ space.
During training, we randomly sample $d$'s
to compute losses which will be introduced later.
Rather than allowing $\bsb{v}_{\text{dir} \rightarrow d}$ to
learn in the whole $\mathcal{W}_0$ space, we leverage
PCA projection to constrain the discovery results in
the informative subspace of $\mathcal{W}$, which will be introduced
in the next subsection in detail.

To conduct the selection along the $\mathcal{W}$-space layer dimension,
we build a branch to predict attention weights on this layer dimension
(depicted by the green boxes in Fig. \ref{fig:navigator} (b) (c)).
Mathematically,
\begin{equation}
    \bsb{v}_{\text{att} \rightarrow d} = \sigma *
    \mathtt{softmax}(\text{att\_logits}_d), \quad d \in 1..m,
\end{equation}
where the term ``$\text{att\_logits}_d$" denotes the output logits of the attention branch
(whose length equal to the $\mathcal{W}$-space layers).
We use a Gaussian kernel $\sigma$ convolution to soften the
outputs of the $\text{softmax}$ function.
Without the convolution operation, the modifications tend to be too sharp
and are likely to cause artifacts on images.
Note that each predicted direction
$\bsb{v}_{\text{dir} \rightarrow d}$ corresponds
to an attention output $\bsb{v}_{\text{att} \rightarrow d}$.
We compute the product of them to obtain the latent modification
, which is then applied to the latent code $w \in \mathcal{W}$:
\begin{align}
    \bsb{v}_{\rightarrow d} &= \bsb{v}_{\text{dir} \rightarrow d} \odot
    \bsb{v}_{\text{att} \rightarrow d}, \\
    \bsb{w}_{\rightarrow d} &= \bsb{w} + \bsb{v}_{\rightarrow d} \quad d \in 1..m, \label{eq:latent_var}
\end{align}
with $\bsb{w}_{\rightarrow d}$ being the edited code 
as illustrated in Fig. \ref{fig:navigator} (b).

\subsection{Informative Subspace in $\mathcal{W}_0$}
\label{sec:info_subspace}
The $\mathcal{W}_0$ space of StyleGAN has been shown highly anisotropic
\cite{hrknen2020ganspace,wang2021aGANGeom},
with a small latent subspace contributing
to almost all generated varieties
(highly informative) while leaving
the rest of the space controlling almost nothing.
This phenomenon hinders our discovery process as the solution
space is sparse.
Concerning this issue, we restrict the
learning of predicted directions $\bsb{v}_{\text{dir}\rightarrow d}$
in the high-informative subspace so that the
navigator can more easily converge to meaningful directions.
In practice, this informative subspace is obtained by
computing PCA on a large number of samples in the $\mathcal{W}_0$ space
similar to \cite{hrknen2020ganspace}, and using the top-$k$ components
(eigenvectors) to define the informative subspace.
(see Appendix Sec. \ref{ap:how_to_pca} for a detailed introduction).

\begin{figure*}[t]
    \begin{center}
       \includegraphics[width=\linewidth]{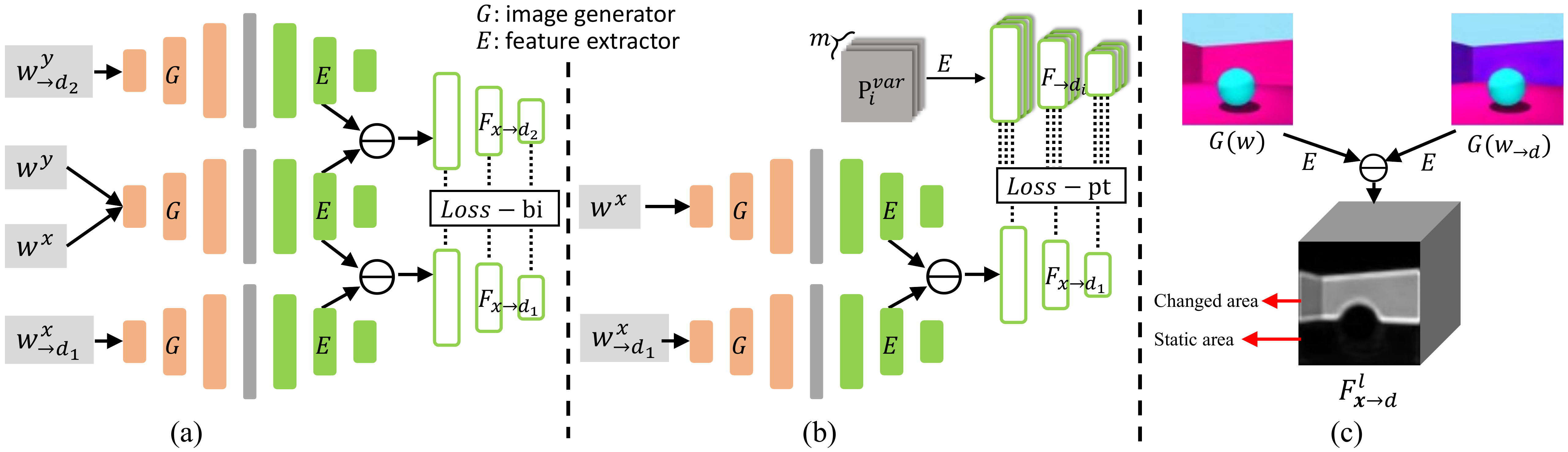}
    \end{center}
        \caption{(a). Binary contrasting process.
        Two directions of change are sampled to contrast with each other.
        (b). Prototype contrasting process.
        One direction of change is sampled to contrast with
        prototype variation patterns.
        These prototype patterns are learned with gradient decent.
        (c). A layer of perceptual difference computed by two images
        differing in a single semantics (wall color).
        The pixel difference is large for changed areas and
        small for static areas.}
    \label{fig:loss}
\end{figure*}

Let the columns of matrix
$V_\text{pca} \in \mathbb{R}^{|\mathcal{W}_0| \times |\mathcal{W}_0|}$
denote the computed principal components
and the left-most $k$ columns form the basis of the used subspace.
Instead of directly learning vectors
$\bsb{v}_{\text{dir} \rightarrow d} \in \mathbb{R}^{|\mathcal{W}_0|}$
in the $\mathcal{W}_0$ space, we learn smaller vectors
$\bsb{v}_{\text{sub}\rightarrow d}$ of
length $k$ and project them into the $\mathcal{W}_0$ space
with the top-$k$ PCA basis:
\begin{equation}
    \bsb{v}_{\text{dir} \rightarrow d} =
    \mathtt{norm}(V_\text{pca}[:, :k] \cdot \bsb{v}_{\text{sub}\rightarrow d}),
    \quad \bsb{v}_{\text{sub}\rightarrow d} \in \mathbb{R}^{k},
\end{equation}
where $\mathtt{norm}$ normalizes the vectors to a constant length.
This projection ensures the learned directions are positioned in the
informative subspace defined by the top-$k$ principal components.
This process is depicted in Fig. \ref{fig:navigator} (c) purple box.


\subsection{Losses}
\label{subsec:losses}
We propose to use pretrained-network features to guide the discovery process.
Since deep features show similar judgment to humans
\cite{zhang2018perceptual}, we hypothesize
this agreement extends to semantic variation structure.
In this paper, we consider losses reflecting
the simple semantic structure of variation consistency
(of the same semantics) and orthogonality (between different semantics).

Given a latent code $\bsb{w} \in \mathcal{W}$ and its modified
version (on direction $d$)
$\bsb{w}_{\rightarrow d} = \bsb{w} + \bsb{v}_{\rightarrow d}$,
the corresponding deep-feature changes are defined as:
\small
\begin{equation}
    [F_{\bsb{x}\rightarrow d}^l]_{l=1..L} =
    (E \circ G)(\bsb{w}_{\rightarrow d}) - (E \circ G)(\bsb{w}),
    1 \leq d \leq m,
\end{equation}
\normalsize
where $G$ is the generator and $E$ is the
feature extractor (e.g. a ConvNet).
Usually the extracted features will contain multiple layers ($l=1..L$).
This variation extraction process is
illustrated in Fig. \ref{fig:loss}
(a) (b) (orange and green layers).

\begin{figure*}[t]
    \centering
   \includegraphics[width=\linewidth]{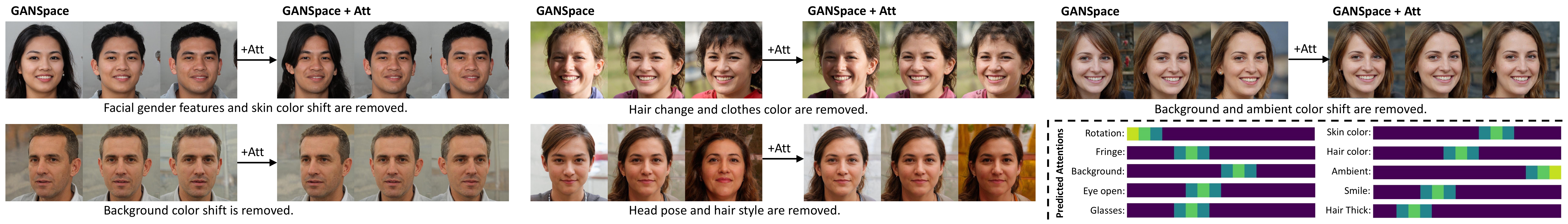}
    \caption{Fixing discovered $\mathcal{W}_0$-space
    directions predicted by Ganspace \cite{hrknen2020ganspace},
    we compare the difference introduced by our proposed attention branch.
    We can see by incorporating the attention module, the interpretability
    of each direction is improved.
    For instance, in the first direction, the facial gender
    features and skin color shift are removed, resulting in
    a specific control on hair length.
    (Bottom-right) Predicted $W$-space layer attentions for
    some semantics.}
\label{fig:evolve}
\end{figure*}

\textbf{Consistency Loss.}
For consistency, we compute two deep-feature changes by applying the same
latent direction $d$ to different latent samples $\bsb{x}$ and $\bsb{y}$,
and force them to be similar to each other:
\begin{equation}
    l^\text{bi}_\text{cons-avg} =
    -\sum_{l=1}^{L}\big[\frac{1}{S}
    \sum_{s=1}^{S}\mathtt{Sim}^2(F_{\bsb{x}\rightarrow d}^{ls},
    F_{\bsb{y}\rightarrow d}^{ls})\big] \label{eq:consistency_avg},
\end{equation}
where $s$ indexes the spatial positions of the feature maps, and
$\texttt{Sim}$ is the cosine similarity function.
Unfortunately, this design has a flaw: it
aggregates scores on all positions evenly but only varied areas
are meaningful to the consistency measurement.
An example of this uneven case is shown in Fig.
\ref{fig:loss} (c) where only the wall color has varied
and the rest areas are static.
Since measuring variation similarity on static areas is meaningless,
we instead adopt an L2-norm mask
of the feature-map difference to implement the weighted aggregation:
\begin{align}
    &l^\text{bi}_{\text{cons}} =
    -\sum_{l=1}^{L}\big[\frac{1}{Q}
    \sum_{s=1}^{S}q(s)\mathtt{Sim}^2(F_{\bsb{x}\rightarrow d}^{ls},
    F_{\bsb{y}\rightarrow d}^{ls})\big], \\
    &q(s) = \frac{\mathtt{nm}(F_{\bsb{x}\rightarrow d}^{s})}
    {\mathtt{nm_{max}}(F_{\bsb{x}\rightarrow d}^{s})} \cdot
    \frac{\mathtt{nm}(F_{\bsb{y}\rightarrow d}^{s})}
    {\mathtt{nm_{max}}(F_{\bsb{y}\rightarrow d}^{s})},
    Q = \sum_{s=1}^{S}q(s),\label{eq:bi-cons}
\end{align}
where $\mathtt{nm}(F_{\bsb{x}\rightarrow d}^{s})$ computes the
L2-norm of the activations on $F_{\bsb{x}\rightarrow d}$ at position $s$.
This weighted aggregation forces the measurement to focus on
the overlapped changes in both inputs.
The division by $Q$ normalizes the weighted aggregation so that
the loss cannot favor large changed areas.
Empirically this version of consistency measurement leads to
better discovery results. We compare different designs in
experiment.

Eq. \ref{eq:bi-cons} compares variations in a
binary manner, where we need to apply the same change to two different codes
(illustrated in Fig. \ref{fig:loss} (a)),
which is not very efficient due to the doubled computation in generator.
To solve this issue, we implement another contrastive learning variant
by learning a set of prototype patterns:
\begin{align}
    &l^\text{pt}_{\text{cons}} = -\sum_{l=1}^{L}\big[\frac{1}{Q}
    \sum_{s=1}^{S}q(s)\mathtt{Sim}^2(F_{\bsb{x}\rightarrow d}^{ls},
    F_{* \rightarrow d}^{ls})\big], \\
    &[F^l_{* \rightarrow d}]_{l=1..L} = E(\mathtt{tanh}(\text{P}^{var}_{d}))
    \label{eq:pt-cons},
\end{align}
where $\text{P}^{var}_{d}\in \mathbb{R}^{h\times w\times c}$
denotes the prototype
pattern of semantic $d$, learned along with the model.
These prototype patterns serve as
hallucinated pictures of the discovered
semantic changes in the RGB space (see visualizations in Fig. \ref{fig:var_patterns}),
which are supposed to cause similar deep-feature
changes as by actual image pairs.

\textbf{Orthogonality Loss.}
Similarly, we define the orthogonality losses (both binary
and prototype implementations), which encourage
feature variations caused by different latent changes to be orthogonal:
\small
\begin{align}
    &l^\text{bi}_\text{orth} =
    \sum_{l=1}^{L}\big[\frac{1}{Q}
    \sum_{s=1}^{S}q(s)\mathtt{Sim}^2(F_{\bsb{x}\rightarrow d}^{ls},
    F_{\bsb{y}\rightarrow d'}^{ls})\big],
    \label{eq:bi-orth}\\
    &l^\text{pt}_{\text{orth}} = \frac{1}{m-1}\sum_{d'\neq
    d}\sum_{l=1}^{L}\big[\frac{1}{Q}
    \sum_{s=1}^{S}q(s)\mathtt{Sim}^2(F_{\bsb{x}\rightarrow d}^{ls},
    F_{* \rightarrow d'}^{ls})\big]
    \label{eq:pt-orth},
\end{align}
\normalsize
where $d,d'$ denote different directions predicted by our navigator.

\textbf{Diversity Regularization.}
We add a small regularization term to encourage the navigator to
proposed different directions in the latent space.
This term practically increases the diversity of discovered concepts:
\begin{equation}
    l_\text{div} = \sum_{i,j} \mathtt{Sim}^2(\bsb{v}_{\text{dir}\rightarrow i},
    \bsb{v}_{\text{dir}\rightarrow j}), \quad i,j \in {1..m}, i \neq j. \label{eq:div}
\end{equation}

In summary, our overall losses (both binary and prototype) are:
\begin{equation}
    l^\text{bi} = l^\text{bi}_\text{cons} +
    l^\text{bi}_\text{orth} + \lambda l_\text{div}
    \quad \text{and} \quad
    l^\text{pt} = l^\text{pt}_\text{cons} +
    l^\text{pt}_\text{orth} + \lambda l_\text{div}
    . \label{eq:final}
\end{equation}

\section{Experiments}
\label{sec:experiments}

\begin{table}[t]
    \begin{center}
        \begin{tabular}{lll}
            \toprule
                Models & $S_\text{disen}$ & $N_\text{discov}$ \\
            \midrule
                Gsp\_dir-non-non & $0.304$ & $9$ \\
                Gsp\_dir-att-bi & $\bsb{0.320}\pm 0.014$ & $10.6\pm 1.02$ \\
                Gsp\_dir-att-pt & $0.327\pm 0.015$ & $\bsb{11.8}\pm 0.75$ \\
            \midrule
                Sefa\_dir-non-non & $0.290$ & $9$ \\
                Sefa\_dir-att-bi & $\bsb{0.328}\pm 0.023$ & $\bsb{13.8}\pm 0.75$ \\
                Sefa\_dir-att-pt & $0.321\pm 0.016$ & $13.0\pm 0.63$ \\
            \bottomrule
        \end{tabular}
    \end{center}
    \caption{Comparing the semantic discovery performance
        with and without our attention module on baselines
        Ganspace \cite{hrknen2020ganspace},
        and Sefa \cite{shen2021closedform}. The shown model names follow
        the pattern: dir\_module - att\_module - loss\_type.}
        \label{table:att_effect}
\end{table}


\subsection{Evaluation Metrics}
\label{subsec:metrics}
We design two quantitative metrics on the
FFHQ dataset \cite{Karras2020ASG} to evaluate the semantic discovery
performance of models.
Considering there are no labeled attributes on this dataset,
we leverage a pretrained state-of-the-art disentanglement model
\cite{Xinqi_cvpr21}
to construct an attribute predictor.
Specifically, for images $X$ generated by the disentangled generator,
we train a predictor $P(X)$ to reconstruct the disentangled
latent codes $\bsb{z} \in R^{p}$, with each dimension
representing a score for an attribute (total of $p$ attributes).

To evaluate a semantic discovery model, we iterate through
all the predicted directions of the model,
apply these modifications to a set of $N$ samples, and compute
their mean absolute difference on the scores predicted
by the pretrained predictor $P$. Mathematically, for $d = 1..m$:
\small
\begin{align}
    A[d, :] = &\frac{A'[d, :]}{\mathtt{max}(A'[d, :])}, \\
    A'[d, :] = \frac{1}{N}\sum_{i=1}^{N}|(P\circ G)(\bsb{w}_i&+
    \bsb{v}_{\rightarrow d})
    - (P\circ G)(\bsb{w}_i)|,
\end{align}
\normalsize
where each entry of
the matrix $A \in R^{m \times p}$ represents
how an attribute score ($i=1..p$) changes on average
when images vary along a discovered semantic direction ($d=1..m$).
Note that this matrix is normalized so
that the maximum value of each row is 1.
Then we define two evaluation quantities:
\begin{align}
    S_\text{disen} = \frac{1}{m} \sum_{d=1}^{m}
    &\mathtt{max}_{1\text{st}}(A[d,:])
    - \mathtt{max}_{2\text{nd}}(A[d,:]), \\
    N_\text{discov} &= \sum_{i=1}^{p}
    \bsb{1}_{\{x| 1 \in x\}}(A[:,i]),
\end{align}
where $\mathtt{max}_{1\text{st}}(A[d,:])$ extracts the largest value
in $A[d,:]$, and $\mathtt{max}_{2\text{nd}}(A[d,:])$ extracts
the second largest value. The $S_\text{disen}$ evaluates how clean a discovered
semantics is because an ideal discovery should 
only cause a single ground-truth attribute to change.
The $N_\text{discov}$ quantity counts the number of semantics discovered by
the model. The term $\bsb{1}_{\{x| 1 \in x\}}(A[:,i])$ is 1 if
the attribute $i$ is at least once predicted as
the maximum value in a row (indicating the
attribute $i$ has been discovered by the model).
Some additional information of evaluation metrics can be
found in Appendix \ref{ap:metrics}.

\subsection{Effectiveness of the Attention Branch}
    In Table \ref{table:att_effect}, we quantitatively compare
    the impact made by our attention branch when applied to Ganspace (Gsp)
    \cite{hrknen2020ganspace} and Sefa \cite{shen2021closedform}
    baselines. The results are obtained
    by training a proposed attention module
    on the top-30 directions predicted by either base models
    in both binary (bi) and prototype-pattern (pt) implementations.
    We can see the attention branch improves the baselines
    on both metrics, especially $N_\text{discov}$.
    This is because the base models predict many directions representing
    overlapped semantics, leading to low total counted number
    of discovered attributes ($N_\text{discov}$), while our attention
    module removes some of these overlapped semantics, leading to
    the emerge of other smaller but diverse attributes.
    In Fig. \ref{fig:evolve}, we qualitatively show the difference
    before and after the use of the attention branch on the
    Ganspace predicted directions. We can see though
    the Ganspace can propose directions to control meaningful
    image changes, many of them are still mixture of multiple semantics.
    In contrast, our learned attentions help to remove some
    correlated changes like the facial gender features, background
    changes, and hair style changes, leading to cleaner semantic controls.

    In Fig. \ref{fig:evolve} bottom-right, we show the predicted attentions
    for some semantic concepts discovered on FFHQ.
    We can see the that high-level changes (e.g. rotation)
    correspond to $\mathcal{W}$-space top layers, facial attribute changes
    (e.g. smile, haircolor) correspond to middle layers,
    and changes like ambient hue correspond to low layers,
    which is similar to results found in \cite{Karras2020ASG}.

\begin{figure*}[t]
    \centering
   \includegraphics[width=\linewidth]{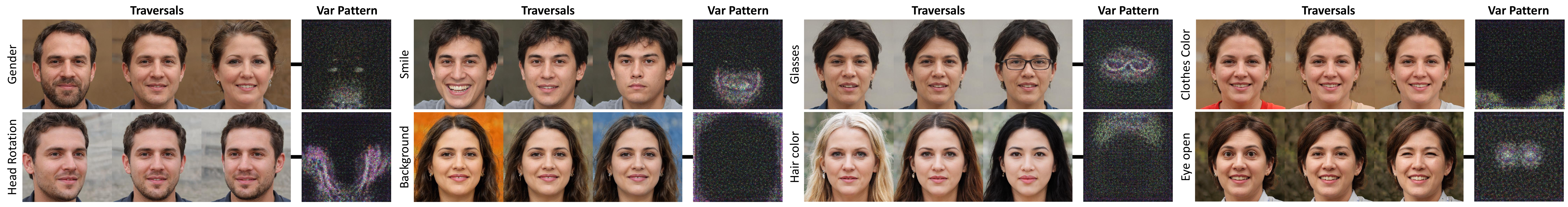}
    \caption{Prototype variation patterns learned
    with our proposed Contra-Pt variant. These patterns learn to
    characterize the changes that are related to the corresponding
    discovered semantics.}
\label{fig:var_patterns}
\end{figure*}

\begin{table}[t]
    \begin{center}
            \begin{tabular}{lll}
                \toprule
                Models & $S_\text{disen}$ & $N_\text{discov}$ \\
                \midrule
                Bi-30 & $0.313\pm 0.015$ & $13.6\pm 0.80$ \\
                Bi-100 & $\bsb{0.357}\pm 0.017$ & $14.2\pm 0.85$ \\
                Bi-300 & $0.340\pm 0.012$ & $\bsb{14.4}\pm 0.80$ \\
                Bi-500 & $0.337\pm 0.014$ & $12.0\pm 0.89$ \\
                \midrule
                Pt-30 & $0.314\pm 0.018$ & $\bsb{14.6}\pm 0.49$ \\
                Pt-100 & $\bsb{0.344}\pm 0.022$ & $14.2\pm 0.75$ \\
                Pt-300 & $0.336\pm 0.013$ & $13.6\pm 0.49$ \\
                Pt-500 & $0.332\pm 0.015$ & $11.8\pm 0.75$ \\
                \bottomrule
        \end{tabular}
    \end{center}
    \caption{Comparing the discovery performance of our models
        learned in PCA subspaces of different sizes.}
    \label{table:k_subspace}
\end{table}

\subsection{Informative Subspace}
    As introduced in Sec. \ref{sec:info_subspace}, there exists
    an informative subspace in the $\mathcal{W}_0$ space that accounts
    for most of the generation variety, and we can restrict the
    discovery process in this dense
    subspace to make the discovery easier.
    In Table \ref{table:k_subspace}, we compare the
    discovery results obtained in subspaces of different sizes.
    We can see when we restrict the subspace to be very small (30-dim),
    the disentanglement score $S_\text{disen}$ becomes the lowest
    in both model variants, while the number of discovered attributes
    $N_\text{discov}$ becomes highest.
    This is because the low-dim subspace is content-dense, thus the model
    can more easily find meaningful changes, leading to higher numer in
    $N_\text{discov}$. However, the semantics discovered in this small subspace
    may be incomplete as some of them are clipped and may be embedded beyond
    this subspace, resulting in partial discovery or artifacts,
    leading to low $S_\text{disen}$.
    On the other hand, using a large subspace (500-dim) leads to lower
    $N_\text{discov}$, meaning it becomes harder to converge to
    meaningful directions as a large part of the space controls null changes.
    In summary, we find that using a subspace
    of a middle size (100 or 300-dim)
    leads to more balanced discovery results.

\subsection{Prototype Variation Patterns}
\label{sec:proto_patterns}
In Fig. \ref{fig:var_patterns}, we present the prototype variation
patterns learned by our Contra-Pt model variant.
We can see the model successfully learns to characterize many
interpretable patterns including some subtle changes like clothes color,
eye opening, and shoulder pose. It is also interesting to see
that the model spontaneously decomposes the semantics spatially
so that they are less like to interfere with each other.
Learning such hallucinated patterns in the RGB space
opens the possibility for conditional discovery
and fine-grained manipulation, as it is possible to constrain the patterns
by some predefined semantic/attribute segmentations.
We leave this interesting extension for future work.


\subsection{Compared to Other Discovery Models}
In Table \ref{table:dis_sota} left, we quantitatively compare
the proposed models
with other existing models
including recognition (Recog) \cite{voynov2020unsupervised}
and LatentCLR (LatCLR) \cite{yuksel2021latentclr}.
We can see:
(1) our proposed contrastive loss performs best among
all models;
(2) existing models can be significantly boosted by incorporating
our proposed $\mathcal{W}$-space attention module.
Besides the performance gain, it is also worth noting that (1)
the Recog model requires to train an extra deep predictor for the discovery
process which takes more training steps to converge
while our models only need to train the navigator (and the
variation patterns in the PT variant); (2) the LatCLR model is designed
to contrast every direction change with each other (an outer product)
in very loss computation, which is highly memory-consuming.
It prevents the model from using features beyond the early few
layers of the generator for its loss computation, limiting the resulted
semantic discovery performance.
In Table. \ref{table:dis_sota} right, we also compare these losses on
disentangled representation learning on 3DShapes dataset
\cite{Kim2018DisentanglingBF}.
We can see our models still outperform the baselines
by an obvious margin.

\begin{table}[t]
    \begin{center}
        \begin{tabular}{lll}
        \toprule
            Model & $S_\text{disen}$ & $N_\text{discov}$ \\
        \midrule
            Recog & $0.249\pm 0.023$ & $4.8\pm 0.75$ \\
            Recog-Att & $0.312\pm 0.015$ & $8.6\pm 0.80$ \\
            LatCLR & $0.267\pm 0.014$ & $9.6\pm 1.02$ \\
            LatCLR-Att & $0.304\pm 0.020$ & $10.2\pm 1.17$ \\
        \midrule
            Contra-Bi & $\bsb{0.357}\pm 0.017$ & $\bsb{14.2}\pm 0.85$ \\
            Contra-Pt & $\bsb{0.344}\pm 0.022$ & $\bsb{14.2}\pm 0.75$ \\
        \bottomrule
        \end{tabular}
    \end{center}
    \caption{Semantic discovery comparison between
    different models with and without the proposed attention
    module on FFHQ dataset \cite{Karras2020ASG}.}
    \label{table:sem_disc_perform}
\end{table}
\begin{table}[t]
    \begin{center}
        \begin{tabular}{llll}
        \toprule
            Model & FVM & MIG & DCI \\
        \midrule
            $\beta$-VAE & $90.2_{\pm 8.1}$ & $39.5_{\pm 8.4}$ & $67.4_{\pm 3.7}$ \\
            FacVAE & $91.4_{\pm 5.0}$ & $39.2_{\pm 8.9}$ & $\bsb{76.3}_{\pm 8.0}$ \\
            Recog-Att & $86.7_{\pm 5.2}$ & $29.1_{\pm 10.2}$ & $49.4_{\pm 8.3}$ \\
            LatCLR-Att & $92.9_{\pm 7.0}$ & $40.4_{\pm 5.3}$ & $70.5_{\pm 12.0}$ \\
        \midrule
            Contra-Bi & $\bsb{93.4}_{\pm 6.2}$ & $\bsb{43.8}_{\pm 7.3}$ & $72.0_{\pm 5.4}$ \\
            Contra-Pt & $\bsb{95.3}_{\pm 4.3}$ & $\bsb{47.2}_{\pm 6.5}$ & $\bsb{74.5}_{\pm 7.0}$ \\
        \bottomrule
        \end{tabular}
    \end{center}
    \caption{Disentangled representation learning comparison on 3DShapes
    dataset \cite{Kim2018DisentanglingBF}.}
    \label{table:dis_sota}
\end{table}

\subsection{Learning Disentangled Representations}
\label{subsec:disen}
With semantics discovered, we can then construct disentangled
representations by learning an encoder that aligns the semantic directions
with axes in a latent space.
This can be achieved by training a model on a dataset composed of
paired images, with each data pair differing in a single semantic direction.
This dataset can be constructed as follows:
\begin{align}
    \{(G(\bsb{w}),
    G(\bsb{w} + \bsb{v}_{d})) \, | \, \bsb{w} \sim p(\bsb{w}),
    d = 1..m\}.
\end{align}
Then we can train an off-the-shelf group-based VAE model
\cite{ijcai2019-348,BouchacourtTN18,pmlr-v119-locatello20a}
to obtain disentangled representations.
The implementation of the group-based VAE is shown in the Appendix Sec.
\ref{ap:group_vae}.
With disentangled representations, we can use quantitative metrics
\cite{Kim2018DisentanglingBF,
Eastwood2018AFF,chen2018isolating} to indirectly evaluate the quality of
discovered semantic directions.

We show the state-of-the-art comparison on 3DShapes in
Table \ref{table:dis_sota}. We can see the derived representations
by our ContraFeat models match or outperform the existing state-of-the-art
disentanglement methods, showing this new pipeline
(discovering semantics + disentangling)
is an effective unsupervised disentangled representation learning approach.
Qualitative results are in Appendix Sec. \ref{ap:3dshapes_trav}.

\begin{figure}[t]
    \centering
   \includegraphics[width=\linewidth]{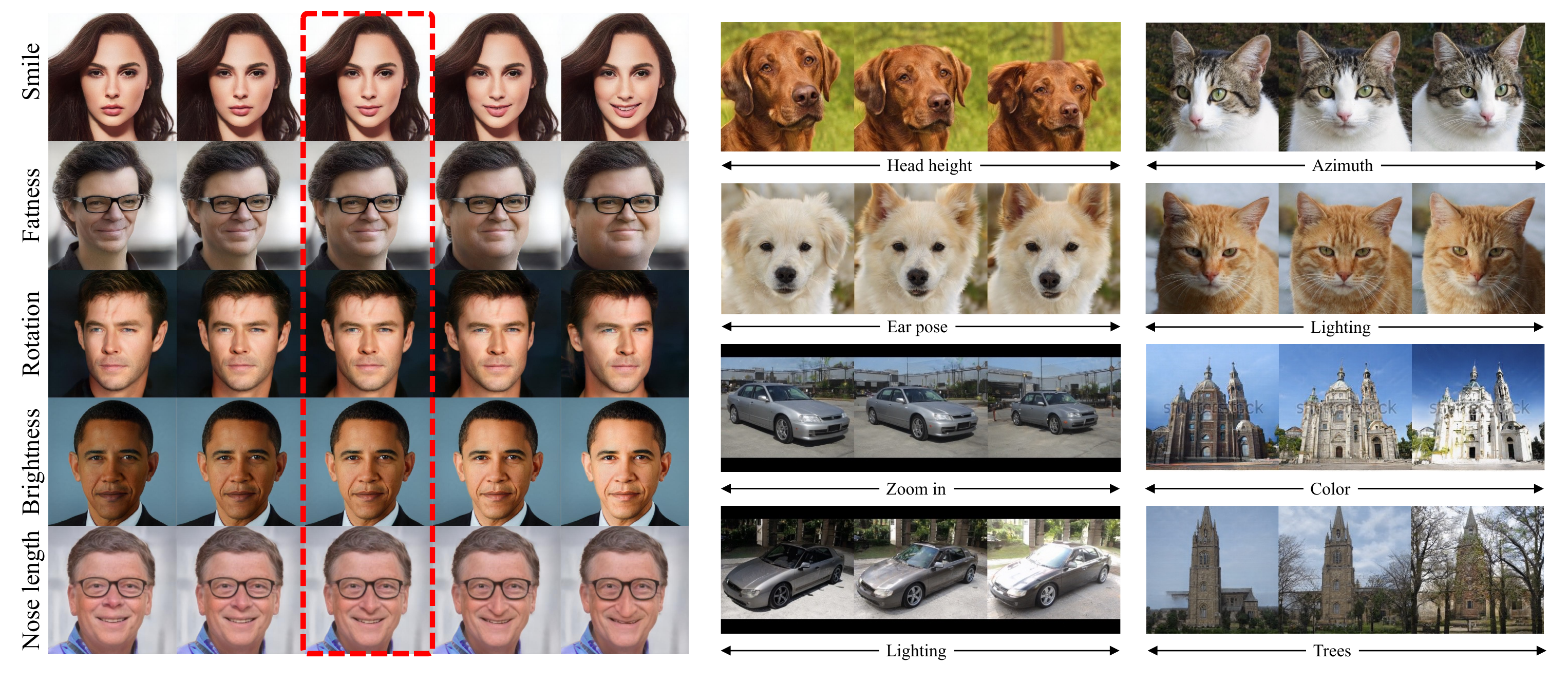}
    \caption{(Left) Some real-image editing results. The red box shows
    the reconstructed original images.
    (Right) Some qualitative discovery results on different datasets.}
\label{fig:more_qualitative}
\end{figure}

\subsection{Real-image Editing and More Qualitative Results}
We can achieve real-image editing by projecting real images
into the $\mathcal{W}$ latent space of StyleGAN.
We adopt the projection method from \cite{Karras2020AnalyzingAI},
with the only modification of learning the projected code
in the extended $\mathcal{W}$ space rather than the
originally-used $\mathcal{W}_0$ space.
After projection, we can apply our discovered semantic directions
to the projected code to edit the image.
In Fig. \ref{fig:more_qualitative} (left), we show some editing results
on real-world images (more results are shown
in the Appendix Sec. \ref{sec:real_image}).
Here we see that we can conduct interesting editing like
fatness and nose length adjustment, which are rarely-seen in
existing other semantic discovery methods.


In Fig. \ref{fig:more_qualitative} right, we show some discovered
semantics on other datasets.
We can see our model can discover specific semantics for different
datasets like the \emph{ear pose} for Dogs dataset and \emph{trees}
for Church dataset, indicating our models can
work adaptively on these datasets.
More discovery results can be found in the Appendix Sec.
\ref{ap:other_sets}.


\section{Conclusion}
\label{sec:conclusion}
Concerning the tedious procedure of manual selection
on $\mathcal{W}$-space layers in existing semantic discovery methods, we
introduced an attention-equipped navigator module to solve this problem.
In order to encourage more effective semantic discovery,
we additionally proposed to restrict the discovery process in
the latent subspace of $\mathcal{W}$ defined by the top-$k$ principal components.
Leveraging the high perceptual capability of deep-network features,
we adopted a pretrained network as a cheap semantic sensor to 
guide the discovery process in a contrastive learning manner by encouraging
consistency and orthogonality in image variations.
Two model variants were introduced with one contrasting variation samples
in a binary way and another contrasting samples with learned
prototype variation patterns.
To quantitatively evaluate semantic discovery models
without ground-truth labels, we designed
two metrics on FFHQ based on a pretrained disentanglement model.
Experimentally, we showed that 
the two proposed models are both effective and
can achieve state-of-the-art semantic discovery performance.
Additionally, we showed that disentangled representations can be
derived from our discovered semantics, and real-image editing can
also be achieved based on image projection techniques.


\section{Acknowledgements}
This work was supported by the Australian Research Council
under projects FL-170100117, IC-190100031, LE-200100049,
DP210101859, the University of Sydney Research Accelerator
(SOAR) Prize, and the University of Sydney 
Faculty of Engineering PhD Completion Award.

%
%
\bibliography{aaai23}

\begin{thebibliography}{46}
\providecommand{\natexlab}[1]{#1}

\bibitem[{Abdal, Qin, and Wonka(2019)}]{9008515}
Abdal, R.; Qin, Y.; and Wonka, P. 2019.
\newblock Image2StyleGAN: How to Embed Images Into the StyleGAN Latent Space?
\newblock In \emph{2019 IEEE/CVF International Conference on Computer Vision
  (ICCV)}, 4431--4440.

\bibitem[{Abdal, Qin, and Wonka(2020)}]{9157575}
Abdal, R.; Qin, Y.; and Wonka, P. 2020.
\newblock Image2StyleGAN++: How to Edit the Embedded Images?
\newblock In \emph{2020 IEEE/CVF Conference on Computer Vision and Pattern
  Recognition (CVPR)}, 8293--8302.

\bibitem[{Bau et~al.(2019)Bau, Zhu, Strobelt, Zhou, Tenenbaum, Freeman, and
  Torralba}]{bau2019gandissect}
Bau, D.; Zhu, J.-Y.; Strobelt, H.; Zhou, B.; Tenenbaum, J.~B.; Freeman, W.~T.;
  and Torralba, A. 2019.
\newblock GAN Dissection: Visualizing and Understanding Generative Adversarial
  Networks.
\newblock In \emph{Proceedings of the International Conference on Learning
  Representations (ICLR)}.

\bibitem[{Bouchacourt, Tomioka, and Nowozin(2018)}]{BouchacourtTN18}
Bouchacourt, D.; Tomioka, R.; and Nowozin, S. 2018.
\newblock Multi-Level Variational Autoencoder: Learning Disentangled
  Representations From Grouped Observations.
\newblock In McIlraith, S.~A.; and Weinberger, K.~Q., eds., \emph{Proceedings
  of the Thirty-Second {AAAI} Conference on Artificial Intelligence},
  2095--2102. {AAAI} Press.

\bibitem[{Brock, Donahue, and Simonyan(2019)}]{Brock2018LargeSG}
Brock, A.; Donahue, J.; and Simonyan, K. 2019.
\newblock Large Scale GAN Training for High Fidelity Natural Image Synthesis.

\bibitem[{Burgess et~al.(2018)Burgess, Higgins, Pal, Matthey, Watters,
  Desjardins, and Lerchner}]{Burgess2018UnderstandingDI}
Burgess, C.~P.; Higgins, I.; Pal, A.; Matthey, L.; Watters, N.; Desjardins, G.;
  and Lerchner, A. 2018.
\newblock Understanding disentangling in beta-VAE.
\newblock \emph{ArXiv}, abs/1804.03599.

\bibitem[{Cao et~al.(2018)Cao, Katzir, Jiang, Lischinski, Cohen-Or, Tu, and
  Li}]{Cao2018DiDADS}
Cao, J.; Katzir, O.; Jiang, P.; Lischinski, D.; Cohen-Or, D.; Tu, C.; and Li,
  Y. 2018.
\newblock DiDA: Disentangled Synthesis for Domain Adaptation.
\newblock \emph{ArXiv}, abs/1805.08019.

\bibitem[{Chen et~al.(2018)Chen, Li, Grosse, and Duvenaud}]{chen2018isolating}
Chen, R. T.~Q.; Li, X.; Grosse, R.; and Duvenaud, D. 2018.
\newblock Isolating Sources of Disentanglement in Variational Autoencoders.
\newblock In \emph{Advances in Neural Information Processing Systems}.

\bibitem[{Chen et~al.(2016)Chen, Duan, Houthooft, Schulman, Sutskever, and
  Abbeel}]{Chen2016InfoGANIR}
Chen, X.; Duan, Y.; Houthooft, R.; Schulman, J.; Sutskever, I.; and Abbeel, P.
  2016.
\newblock InfoGAN: Interpretable Representation Learning by Information
  Maximizing Generative Adversarial Nets.
\newblock In \emph{NIPS}.

\bibitem[{Creager et~al.(2019)Creager, Madras, Jacobsen, Weis, Swersky,
  Pitassi, and Zemel}]{Creager2019FlexiblyFR}
Creager, E.; Madras, D.; Jacobsen, J.-H.; Weis, M.~A.; Swersky, K.; Pitassi,
  T.; and Zemel, R.~S. 2019.
\newblock Flexibly Fair Representation Learning by Disentanglement.
\newblock In \emph{ICML}.

\bibitem[{Dosovitskiy, Springenberg, and
  Brox(2014)}]{Dosovitskiy2014LearningTG}
Dosovitskiy, A.; Springenberg, J.~T.; and Brox, T. 2014.
\newblock Learning to generate chairs with convolutional neural networks.
\newblock \emph{2015 IEEE Conference on Computer Vision and Pattern Recognition
  (CVPR)}, 1538--1546.

\bibitem[{Eastwood and Williams(2018)}]{Eastwood2018AFF}
Eastwood, C.; and Williams, C. K.~I. 2018.
\newblock A Framework for the Quantitative Evaluation of Disentangled
  Representations.
\newblock In \emph{ICLR}.

\bibitem[{Goodfellow et~al.(2014)Goodfellow, Pouget-Abadie, Mirza, Xu,
  Warde-Farley, Ozair, Courville, and Bengio}]{Goodfellow2014GenerativeAN}
Goodfellow, I.~J.; Pouget-Abadie, J.; Mirza, M.; Xu, B.; Warde-Farley, D.;
  Ozair, S.; Courville, A.~C.; and Bengio, Y. 2014.
\newblock Generative Adversarial Networks.
\newblock In \emph{NIPS}.

\bibitem[{Higgins et~al.(2017)Higgins, Matthey, Pal, Burgess, Glorot,
  Botvinick, Mohamed, and Lerchner}]{Higgins2017betaVAELB}
Higgins, I.; Matthey, L.; Pal, A.; Burgess, C.; Glorot, X.; Botvinick, M.~M.;
  Mohamed, S.; and Lerchner, A. 2017.
\newblock beta-VAE: Learning Basic Visual Concepts with a Constrained
  Variational Framework.
\newblock In \emph{ICLR}.

\bibitem[{Hosoya(2019)}]{ijcai2019-348}
Hosoya, H. 2019.
\newblock Group-based Learning of Disentangled Representations with
  Generalizability for Novel Contents.
\newblock In \emph{Proceedings of the Twenty-Eighth International Joint
  Conference on Artificial Intelligence, {IJCAI-19}}, 2506--2513. International
  Joint Conferences on Artificial Intelligence Organization.

\bibitem[{Härkönen et~al.(2020)Härkönen, Hertzmann, Lehtinen, and
  Paris}]{hrknen2020ganspace}
Härkönen, E.; Hertzmann, A.; Lehtinen, J.; and Paris, S. 2020.
\newblock GANSpace: Discovering Interpretable GAN Controls.
\newblock In \emph{Proc. NeurIPS}.

\bibitem[{Jahanian, Chai, and Isola(2020)}]{gansteerability}
Jahanian, A.; Chai, L.; and Isola, P. 2020.
\newblock On the "steerability" of generative adversarial networks.
\newblock In \emph{International Conference on Learning Representations}.

\bibitem[{Jeon, Lee, and Kim(2018)}]{Jeon2018IBGANDR}
Jeon, I.; Lee, W.; and Kim, G. 2018.
\newblock IB-GAN: Disentangled Representation Learning with Information
  Bottleneck GAN.
\newblock \emph{ArXiv}.

\bibitem[{Karras et~al.(2018)Karras, Aila, Laine, and
  Lehtinen}]{Karras2017ProgressiveGO}
Karras, T.; Aila, T.; Laine, S.; and Lehtinen, J. 2018.
\newblock Progressive Growing of GANs for Improved Quality, Stability, and
  Variation.

\bibitem[{Karras, Laine, and Aila(2020)}]{Karras2020ASG}
Karras, T.; Laine, S.; and Aila, T. 2020.
\newblock A Style-Based Generator Architecture for Generative Adversarial
  Networks.
\newblock \emph{IEEE transactions on pattern analysis and machine
  intelligence}.

\bibitem[{Karras et~al.(2020)Karras, Laine, Aittala, Hellsten, Lehtinen, and
  Aila}]{Karras2020AnalyzingAI}
Karras, T.; Laine, S.; Aittala, M.; Hellsten, J.; Lehtinen, J.; and Aila, T.
  2020.
\newblock Analyzing and Improving the Image Quality of StyleGAN.
\newblock \emph{CVPR}.

\bibitem[{Khemakhem et~al.(2020)Khemakhem, Kingma, Monti, and
  Hyvarinen}]{pmlr-v108-khemakhem20a}
Khemakhem, I.; Kingma, D.; Monti, R.; and Hyvarinen, A. 2020.
\newblock Variational Autoencoders and Nonlinear ICA: A Unifying Framework.
\newblock In Chiappa, S.; and Calandra, R., eds., \emph{Proceedings of the
  Twenty Third International Conference on Artificial Intelligence and
  Statistics}, volume 108 of \emph{Proceedings of Machine Learning Research},
  2207--2217. PMLR.

\bibitem[{Kim and Mnih(2018)}]{Kim2018DisentanglingBF}
Kim, H.; and Mnih, A. 2018.
\newblock Disentangling by Factorising.
\newblock In \emph{ICML}.

\bibitem[{Kingma et~al.(2014)Kingma, Mohamed, Rezende, and
  Welling}]{Kingma2014SemisupervisedLW}
Kingma, D.~P.; Mohamed, S.; Rezende, D.~J.; and Welling, M. 2014.
\newblock Semi-supervised Learning with Deep Generative Models.
\newblock In \emph{NIPS}.

\bibitem[{Kulkarni et~al.(2015)Kulkarni, Whitney, Kohli, and
  Tenenbaum}]{Kulkarni2015DeepCI}
Kulkarni, T.~D.; Whitney, W.~F.; Kohli, P.; and Tenenbaum, J.~B. 2015.
\newblock Deep Convolutional Inverse Graphics Network.
\newblock In \emph{NIPS}.

\bibitem[{Kumar, Sattigeri, and Balakrishnan(2018)}]{Kumar2017VariationalIO}
Kumar, A.; Sattigeri, P.; and Balakrishnan, A. 2018.
\newblock Variational Inference of Disentangled Latent Concepts from Unlabeled
  Observations.
\newblock In \emph{ICLR}.

\bibitem[{Lample et~al.(2017)Lample, Zeghidour, Usunier, Bordes, Denoyer, and
  Ranzato}]{Lample2017FaderNM}
Lample, G.; Zeghidour, N.; Usunier, N.; Bordes, A.; Denoyer, L.; and Ranzato,
  M. 2017.
\newblock Fader Networks: Manipulating Images by Sliding Attributes.
\newblock In \emph{NIPS}, volume abs/1706.00409.

\bibitem[{Lee et~al.(2018)Lee, Tseng, Huang, Singh, and
  Yang}]{Lee2018DiverseIT}
Lee, H.-Y.; Tseng, H.-Y.; Huang, J.-B.; Singh, M.~K.; and Yang, M.-H. 2018.
\newblock Diverse Image-to-Image Translation via Disentangled Representations.
\newblock \emph{ECCV}, abs/1808.00948.

\bibitem[{Lin et~al.(2020)Lin, Thekumparampil, Fanti, and
  Oh}]{Lin2019InfoGANCRDG}
Lin, Z.; Thekumparampil, K.~K.; Fanti, G.; and Oh, S. 2020.
\newblock InfoGAN-CR: Disentangling Generative Adversarial Networks with
  Contrastive Regularizers.
\newblock \emph{ICML}.

\bibitem[{Locatello et~al.(2019{\natexlab{a}})Locatello, Abbati, Rainforth,
  Bauer, Sch{\"o}lkopf, and Bachem}]{Locatello2019OnTF}
Locatello, F.; Abbati, G.; Rainforth, T.; Bauer, S.; Sch{\"o}lkopf, B.; and
  Bachem, O. 2019{\natexlab{a}}.
\newblock On the Fairness of Disentangled Representations.
\newblock In \emph{NeurIPS}.

\bibitem[{Locatello et~al.(2019{\natexlab{b}})Locatello, Bauer, Lucic,
  R{\"a}tsch, Gelly, Sch{\"o}lkopf, and Bachem}]{Locatello2018ChallengingCA}
Locatello, F.; Bauer, S.; Lucic, M.; R{\"a}tsch, G.; Gelly, S.; Sch{\"o}lkopf,
  B.; and Bachem, O. 2019{\natexlab{b}}.
\newblock Challenging Common Assumptions in the Unsupervised Learning of
  Disentangled Representations.
\newblock In \emph{ICML}.

\bibitem[{Locatello et~al.(2020)Locatello, Poole, Raetsch, Sch{\"o}lkopf,
  Bachem, and Tschannen}]{pmlr-v119-locatello20a}
Locatello, F.; Poole, B.; Raetsch, G.; Sch{\"o}lkopf, B.; Bachem, O.; and
  Tschannen, M. 2020.
\newblock Weakly-Supervised Disentanglement Without Compromises.
\newblock In III, H.~D.; and Singh, A., eds., \emph{Proceedings of the 37th
  International Conference on Machine Learning}, volume 119 of
  \emph{Proceedings of Machine Learning Research}, 6348--6359. PMLR.

\bibitem[{Peng et~al.(2019)Peng, Huang, Sun, and Saenko}]{Peng2019DomainAL}
Peng, X.; Huang, Z.; Sun, X.; and Saenko, K. 2019.
\newblock Domain Agnostic Learning with Disentangled Representations.
\newblock In \emph{ICML}.

\bibitem[{Plumerault, Borgne, and Hudelot(2020)}]{Plumerault2020Controlling}
Plumerault, A.; Borgne, H.~L.; and Hudelot, C. 2020.
\newblock Controlling generative models with continuous factors of variations.
\newblock In \emph{International Conference on Learning Representations}.

\bibitem[{Radford et~al.(2021)Radford, Kim, Hallacy, Ramesh, Goh, Agarwal,
  Sastry, Askell, Mishkin, Clark, Krueger, and Sutskever}]{CLIPmodel}
Radford, A.; Kim, J.~W.; Hallacy, C.; Ramesh, A.; Goh, G.; Agarwal, S.; Sastry,
  G.; Askell, A.; Mishkin, P.; Clark, J.; Krueger, G.; and Sutskever, I. 2021.
\newblock Learning Transferable Visual Models From Natural Language
  Supervision.
\newblock \emph{ICML}.

\bibitem[{Shen et~al.(2019)Shen, Gu, Tang, and Zhou}]{Shen2019InterpretingTL}
Shen, Y.; Gu, J.; Tang, X.; and Zhou, B. 2019.
\newblock Interpreting the Latent Space of GANs for Semantic Face Editing.
\newblock \emph{ArXiv}, abs/1907.10786.

\bibitem[{Shen and Zhou(2021)}]{shen2021closedform}
Shen, Y.; and Zhou, B. 2021.
\newblock Closed-Form Factorization of Latent Semantics in GANs.
\newblock In \emph{CVPR}.

\bibitem[{Spingarn, Banner, and Michaeli(2021)}]{spingarn2021gan}
Spingarn, N.; Banner, R.; and Michaeli, T. 2021.
\newblock {\{}GAN{\}} ''Steerability'' without optimization.
\newblock In \emph{International Conference on Learning Representations}.

\bibitem[{Voynov and Babenko(2020)}]{voynov2020unsupervised}
Voynov, A.; and Babenko, A. 2020.
\newblock Unsupervised discovery of interpretable directions in the gan latent
  space.
\newblock In \emph{International Conference on Machine Learning}, 9786--9796.
  PMLR.

\bibitem[{Wang and Ponce(2021)}]{wang2021aGANGeom}
Wang, B.; and Ponce, C.~R. 2021.
\newblock A Geometric Analysis of Deep Generative Image Models and Its
  Applications.
\newblock In \emph{International Conference on Learning Representations}.

\bibitem[{Xing et~al.(2019)Xing, Han, Gao, Zhu, and
  Wu}]{Xing2019UnsupervisedDO}
Xing, X.; Han, T.; Gao, R.; Zhu, S.-C.; and Wu, Y.~N. 2019.
\newblock Unsupervised Disentangling of Appearance and Geometry by Deformable
  Generator Network.
\newblock \emph{2019 IEEE/CVF Conference on Computer Vision and Pattern
  Recognition (CVPR)}, 10346--10355.

\bibitem[{Yan et~al.(2016)Yan, Yang, Sohn, and Lee}]{Yan2016Attribute2ImageCI}
Yan, X.; Yang, J.; Sohn, K.; and Lee, H. 2016.
\newblock Attribute2Image: Conditional Image Generation from Visual Attributes.
\newblock In \emph{ECCV}.

\bibitem[{Y{\"u}ksel et~al.(2021)Y{\"u}ksel, Simsar, Er, and
  Yanardag}]{yuksel2021latentclr}
Y{\"u}ksel, O.~K.; Simsar, E.; Er, E.~G.; and Yanardag, P. 2021.
\newblock LatentCLR: A Contrastive Learning Approach for Unsupervised Discovery
  of Interpretable Directions.
\newblock In \emph{Proceedings of the IEEE/CVF International Conference on
  Computer Vision (ICCV)}, 14263--14272.

\bibitem[{Zhang et~al.(2018)Zhang, Isola, Efros, Shechtman, and
  Wang}]{zhang2018perceptual}
Zhang, R.; Isola, P.; Efros, A.~A.; Shechtman, E.; and Wang, O. 2018.
\newblock The Unreasonable Effectiveness of Deep Features as a Perceptual
  Metric.
\newblock In \emph{CVPR}.

\bibitem[{Zhu, Xu, and Tao(2021{\natexlab{a}})}]{Xinqi_liegroupvae_icml21}
Zhu, X.; Xu, C.; and Tao, D. 2021{\natexlab{a}}.
\newblock Commutative Lie Group VAE for Disentanglement Learning.
\newblock In \emph{ICML}.

\bibitem[{Zhu, Xu, and Tao(2021{\natexlab{b}})}]{Xinqi_cvpr21}
Zhu, X.; Xu, C.; and Tao, D. 2021{\natexlab{b}}.
\newblock Where and What? Examining Interpretable Disentangled Representations.
\newblock In \emph{CVPR}.

\end{thebibliography}

\clearpage

\appendix

\section{How to Compute Informative Subspace}
\label{ap:how_to_pca}
We follow the procedure in \cite{hrknen2020ganspace} to compute the
principal components in the $\mathcal{W}_0$ latent space of StyleGANs.
Specifically, we first sample 1,000,000 points in the isotropic latent
space $Z$ of StyleGAN2: $\bsb{z}_i \sim \mathcal{N}(0, 1)$,
and compute the corresponding intermediate latent codes
in the $\mathcal{W}_0$ space:
\begin{equation}
    \{\bsb{w}_i | \bsb{w}_i = G_{map}(\bsb{z}_i),
    \bsb{z}_i \sim \mathcal{N}(0, 1)\}.
\end{equation}
We conduct PCA on  these vectors, and obtain the PCA matrix
$V_\text{pca}$ composed of eigenvectors which serve as the new basis of
the $\mathcal{W}_0$ latent space.
These eigenvectors are ordered by eigenvalues from high to low,
revealing the ranked informativeness significance in the latent space.
To obtain an informative subspace, we use
the first few ($k$) largest eigenvectors from the
PCA matrix $V_\text{pca}[:, :k]$ as the subspace basis
to support the space projection process described in Sec.
\ref{sec:info_subspace}.

\section{Group-based VAE}
\label{ap:group_vae}
After obtaining a set of predicted semantic directions, we train
a group-based VAE to distill the knowledge to a disentangled representation.
We generate a dataset of image-pairs using the discovered directions:
\begin{equation}
    \{(G(\bsb{w}),
    G(\bsb{w} + \bsb{v}_{d})) \, | \, \bsb{w} \sim p(\bsb{w}),
    d = 1..m\}.
\end{equation}
This dataset consists of paired samples with each sample differing
in only a single semantic concept.

A group-based VAE loss is similar to a regular VAE loss
but has different operations
on the shared dimensions and non-shared dimensions for image pairs.
The posteriors of image pairs are computed as (we follow the formulation
in \cite{pmlr-v119-locatello20a}):
\small
\begin{equation}
    \hat{q}(\hat{\bsb{z}}_{i}|\bsb{x}_1) =
    \begin{cases}
    \mathtt{avg}(
    q(\bsb{z}_i|\bsb{x}_1), q(\bsb{z}_i|\bsb{x}_2)),
    & \text{if $\bsb{z}_i$ is shared,} \\
    q(\bsb{z}_i|\bsb{x}_1),
    & \text{otherwise}.
    \end{cases}\label{eq:gvae}
\end{equation}
\normalsize
where $q(\bsb{z}|\bsb{x})$ is the encoder in a regular VAE which predicts the
mean and variance of the embedding.
The $\mathtt{avg}$ function takes the average of two values.
Eq. \ref{eq:gvae} states that, when we predict the embedding of an image
$\hat{q}(\bsb{z}|\bsb{x})$, instead of only considering the output of
itself $\bsb{x}_1$, we also take into account the output from its paired
sample $\bsb{x}_2$ to form their shared dimension prediction
by averaging: $\mathtt{avg}(q(\bsb{z}_i|\bsb{x}_1), q(\bsb{z}_i|\bsb{x}_2))$.
This forces the two samples to have the same latent prediction on the shared
semantics, leading to the disentanglement of this shared semantic dimension
from other dimensions.

The loss objective of a group-based VAE is defined as:
\small
\begin{align}
    \text{max}\ \mathbb{E}_{(\bsb{x_1}, \bsb{x_2})}&\mathbb{E}_{
        \hat{q}(\hat{\bsb{z}}|\bsb{x}_1)}
    \text{log}(p(\bsb{x}_1|\hat{\bsb{z}}))
    + \mathbb{E}_{\hat{q}(\hat{\bsb{z}}|\bsb{x}_2)}
    \text{log}(p(\bsb{x}_2|\hat{\bsb{z}})) \nonumber \\
    &- D_{KL}(\hat{q}(\hat{\bsb{z}}|\bsb{x}_1)||p(\hat{\bsb{z}}))
    - D_{KL}(\hat{q}(\hat{\bsb{z}}|\bsb{x}_2)||p(\hat{\bsb{z}})).
\end{align}
\normalsize
It will result in a model consisting of an
encoder $q(\bsb{z}|\bsb{x})$ and a decoder $p(\bsb{x}|\bsb{z})$.
The encoder $q(\bsb{z}|\bsb{x})$ extracts the representation (latent code)
of an image, and the decoder $p(\bsb{x}|\bsb{z})$ synthesize an
image based on a latent code.

\section{Additional Information of Evaluation Metrics}
\label{ap:metrics}
The attributes used in evaluation are: ['azimuth', 'haircolor', 'smile',
'gender', 'main\_fringe', 'left\_fringe', 'age', 'light\_right', 'light\_left',
'light\_vertical', 'hair\_style', 'clothes\_color', 'saturation', 'ambient\_color',
'elevation', 'neck', 'right\_shoulder', 'left\_shoulder', 'backgrounds',
'right\_object', 'left\_object']. These are the attributes from \cite{Xinqi_cvpr21}.
The quantity $S_\text{disen}$ works similarly to \emph{precision}
as it describes the quality of the predicted semantics,
while the $N_\text{discov}$ quantity is similar to
\emph{recall} as it describes
how many ground-truth attributes are discovered.
Note that the usefulness of the proposed metrics are bottlenecked by the
the used attribute predictor. The more faithfully the
predictor reflects the semantic structure inside a dataset,
the more accurate the metrics are.
For the predictor on FFHQ used in our experiments, major attributes
like azimuth, gender, smiling and head elevation can be
evaluated, while more subtle attributes like eye opening, nose length,
and face length can not.

\section{Ablation Study on the Diversity Term}
\label{ap:ablation_div}
We show the comparison results of different diversity constraints $\lambda$
in Table \ref{table:div_reg_bi} \ref{table:div_reg_pt}
on both of our proposed variants (Bi and Pt).
We can see when we deactivate the diversity constraint ($\lambda=0$)
the disentanglement score $S_\text{disen}$ is relatively low,
and the number of discovered semantics $N_\text{discov}$ is also low.
This is caused by that the model tends to propose duplicated semantic
changes without this diversity control. When $\lambda$ is too high,
the model performance becomes the worst. This is because the too strong
constraint loss dominates our main loss, and the model only tries to
find some directions that are orthogonal in the $\mathcal{W}$ space
rather than leveraging the perceptual power of the deep features
to obtain perceptually different semantics.
A more balanced constraint term comes at around $\lambda=0.01$ and $0.1$
and we use $\lambda=0.01$ in our other experiments.
\begin{table}[t]
\begin{center}
    \begin{tabular}{lll}
        \toprule
            Models & $S_\text{disen}$ & $N_\text{discov}$ \\
        \midrule
            Pt-$\lambda =0$ & $0.349\pm 0.024$ & $11.8\pm 0.75$ \\
            Pt-$\lambda = 0.0001$ & $0.354\pm 0.017$ & $14.2\pm 0.80$ \\
            Pt-$\lambda = 0.001$ & $0.350\pm 0.020$ & $13.0\pm 1.12$ \\
            Pt-$\lambda = 0.01$ & $\bsb{0.377}\pm 0.015$ & $13.6\pm 0.80$ \\
            Pt-$\lambda = 0.1$ & $0.362\pm 0.024$ & $14.1\pm 0.63$ \\
            Pt-$\lambda = 1$ & $0.356\pm 0.021$ & $11.3\pm 1.00$ \\
            Pt-$\lambda = 10$ & $0.331\pm 0.011$ & $10.8\pm 1.25$ \\
            Pt-$\lambda = 100$ & $0.334\pm 0.019$ & $10.5\pm 1.12$ \\
        \bottomrule
    \end{tabular}
\end{center}
\caption{Comparing different diversity constraints $\lambda$
on Bi variant.}
\label{table:div_reg_bi}
\end{table}
\begin{table}[t]
    \begin{center}
        \begin{tabular}{lll}
            \toprule
            Models & $S_\text{disen}$ & $N_\text{discov}$ \\
            \midrule
            Bi-$\lambda =0$ & $0.341\pm 0.016$ & $9.8\pm 1.08$ \\
            Bi-$\lambda = 0.0001$ & $0.338\pm 0.025$ & $11.8\pm 1.12$ \\
            Bi-$\lambda = 0.001$ & $0.335\pm 0.023$ & $11.1\pm 1.30$ \\
            Bi-$\lambda = 0.01$ & $\bsb{0.357}\pm 0.018$ & $14.2\pm 1.00$ \\
            Bi-$\lambda = 0.1$ & $\bsb{0.361}\pm 0.013$ & $13.5\pm 1.42$ \\
            Bi-$\lambda = 1$ & $0.328\pm 0.022$ & $13.6\pm 0.85$ \\
            Bi-$\lambda = 10$ & $0.322\pm 0.019$ & $12.0\pm 1.14$ \\
            Bi-$\lambda = 100$ & $0.322\pm 0.014$ & $11.8\pm 0.89$ \\
            \bottomrule
        \end{tabular}
    \end{center}
    \caption{Comparing different diversity constraints $\lambda$
    on Pt variant.}
\label{table:div_reg_pt}
\end{table}

\begin{table}[t]
    \begin{center}
        \begin{tabular}{llll}
        \toprule
            Model & Pooled & No-Foc & L2-Mask \\
        \midrule
            Contra-Bi & 0.266 (9.8) & 0.314 (11.8) & \textbf{0.357 (14.2)} \\
            Contra-Pt & 0.311 (11.0) & 0.298 (11.6) & \textbf{0.344 (14.2)} \\
        \bottomrule
        \end{tabular}
    \end{center}
    \caption{Discovery performance comparison between models
    with different losses ($N_\text{discov}$ show in parentheses).}
    \label{table:bi_vs_pt}
\end{table}

\section{Spatial L2-Mask}
\label{sec:avg_vs_masked}
We compare our spatial L2-mask design in our losses with two baselines:
average-pooled features instead of feature maps (Pooled), and
Eq. \ref{eq:consistency_avg} without varied-area focus (No-Foc).
We show the comparison results in Table \ref{table:bi_vs_pt},
evaluated with $S_\text{disen}$ and $N_\text{discov}$ (in parentheses).
We can see our L2-Mask design outperforms the two baselines by
an obvious margin.
It implies that the spatial awareness in discovering semantics
in the image domain is important
as most interpretable concepts are in fact locally identifiable.
To seek a more detailed analysis of the improvement,
we compare the different losses on 3DShapes dataset \cite{Kim2018DisentanglingBF}
under different synthetic settings with controlled ground-truth factors. 
The detailed experimental setup 
and results are presented in 
Sec. \ref{ap:l2_mask_compare} and Sec. \ref{ap:l2_mask_compare_results} respectively.

\begin{table}[t]
    \begin{center}
        \begin{tabular}{llll}
        \toprule
            Setting & Pooled & No-Foc & L2-Mask \\
        \midrule
            (a) Pure & -0.3509 & -0.2378 & \textbf{-0.3977} \\
            (b) Mixed & -0.0708 & -0.0577 & -0.0656 \\
        \bottomrule
        \end{tabular}
    \end{center}
    \caption{Three types of losses obtained in two latent-space settings.
    See Sec. \ref{sec:avg_vs_masked} for more information.}
    \label{table:l2_mask_compare}
\end{table}

\section{L2-Mask Experiment Setting Introduction}
\label{ap:l2_mask_compare}
To seek a more detailed analysis of the improvement,
we compute different losses on 3DShapes \cite{Kim2018DisentanglingBF}
synthetic settings:
(a) in a 2-D latent space, the two latent directions (axes) each controlling
a disentangled semantics, and (b) a rotated version of setting (a) so
that each direction controlling a linear combination of the two semantics.
To obtain these two settings, we first train a (nearly) perfect disentangled VAE model
on 3DShapes dataset, and use its decoder to generate paired
samples to conduct this experiment. Since the decoder
is already disentangled and can represent each semantic variation (including
object shape, object size, object color, wall color, floor color,
and orientation) by a single latent dimension, we can generate
(1) an image pair varying in a single semantics by perturbing the corresponding
latent dimension, and (2) image pairs varying in
two semantics by changing the corresponding two
latent dimensions at the same time.

In this experiment, we compare different losses
under the two different settings of semantic entanglement.
The first setting is computing losses on disentangled semantic pairs.
For consistency loss, we apply the same latent dimension change
to two different codes (sampled from the latent space of the VAE decoder).
For orthogonality loss, we apply different latent dimension changes
to the two codes.
The second setting is entangled semantic pairs.
Instead of changing only a single dimension of the decoder,
we simultaneously change two dimensions (two semantic concepts)
to simulate the change of composed semantics:
\begin{equation}
    l_\text{disen} = l_\text{cons}(\bsb{z}^{1}_{\rightarrow \bsb{a}},
    \bsb{z}^{2}_{\rightarrow \bsb{a}})
    + l_\text{orth}(\bsb{z}^{1}_{\rightarrow \bsb{a}},
    \bsb{z}^{2}_{\rightarrow \bsb{b}}).
\end{equation}
Same as the first setting, we can compute our designed losses
with these composed semantic changes.
Note that in the second setting, we compute the orthogonality losses
with directions changing in $\bsb{a}+\bsb{b}$ and $\bsb{a}-\bsb{b}$,
with $\bsb{a}$ and $\bsb{b}$ being two latent-axis directions of the decoder.
These constructed directions simulate
the linearly-combined (entangled) semantics (a+b and a-b):
\begin{equation}
    l_\text{en} = l_\text{cons}(\bsb{z}^{1}_{\rightarrow \bsb{a}+\bsb{b}},
    \bsb{z}^{2}_{\rightarrow \bsb{a} + \bsb{b}})
    + l_\text{orth}(\bsb{z}^{1}_{\rightarrow \bsb{a}+\bsb{b}},
    \bsb{z}^{2}_{\rightarrow \bsb{a}-\bsb{b}}).
\end{equation}

For each entry in the results (Table \ref{table:l2_mask_compare}),
we compute the losses obtained
on all semantic pairs (e.g. wall color vs floor color,
object shape vs wall color, etc.) on 3DShapes dataset.
When computing on each semantic pair, we average the losses over
1000 samples. The final score is obtained by averaging
the losses obtained from all combination of semantic pairs.

\section{L2-Mask Experiment Results on 3DShapes}
\label{ap:l2_mask_compare_results}
In Table. \ref{table:l2_mask_compare} we show the results
in different settings,
with each entry containing the averaged loss value
of all ground-truth factor-pair combinations on
3DShapes (each factor-pair being averaged with 1000 samples).
We can see: (1) All the three losses show obvious
decrease in the disentangled setting (a)
compared to the mixed setting (b), indicating all losses have
the potential to encourage the directions to converge
to the disentangled results.
This is because clean semantics can obtain better consistency
and orthogonality scores in general.
(2) While all the losses show similar values in the mixed setting,
our L2-Mask design obtains the lowest quantity in the disentangled setting.
This indicates that our L2-Mask design can more favorably identify the
disentangled directions than the other two, and thus potentially
leads to more effective semantic discovery through training.
Note that this experiment shows an example of the unidentifiability problem
discussed in \cite{Locatello2018ChallengingCA} and
\cite{pmlr-v108-khemakhem20a}, and the results indicate that
leveraging spatial biases in images
can alleviate this unidentifiability
problem on the demonstrated dataset.

\section{More Real-image Editing}
\label{sec:real_image}
\begin{figure}
\begin{center}
   \includegraphics[width=\linewidth]{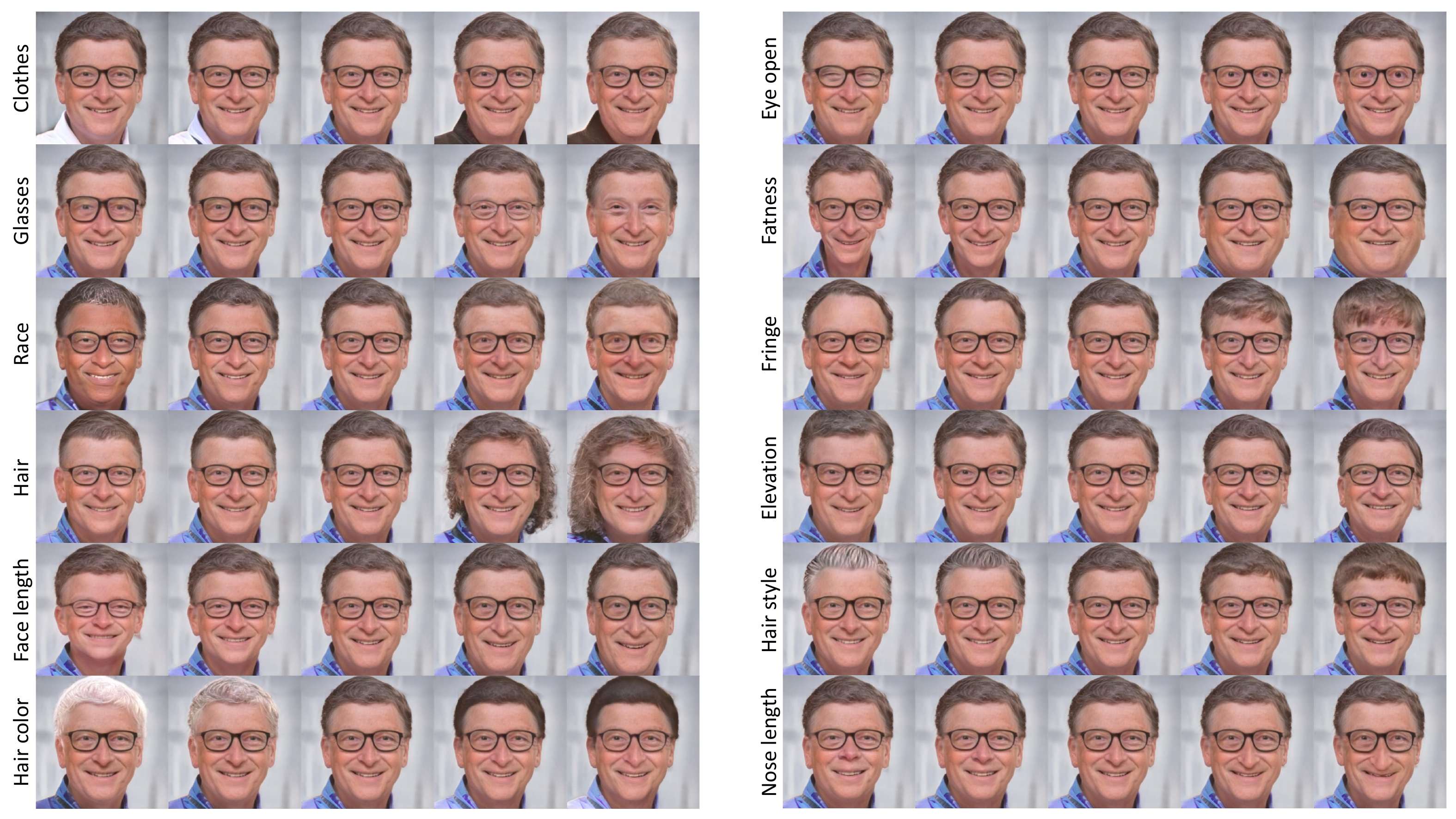}
\end{center}
    \caption{Real-image edit 1.}
\label{fig:bill_gates}
\end{figure}
\begin{figure}
\begin{center}
   \includegraphics[width=\linewidth]{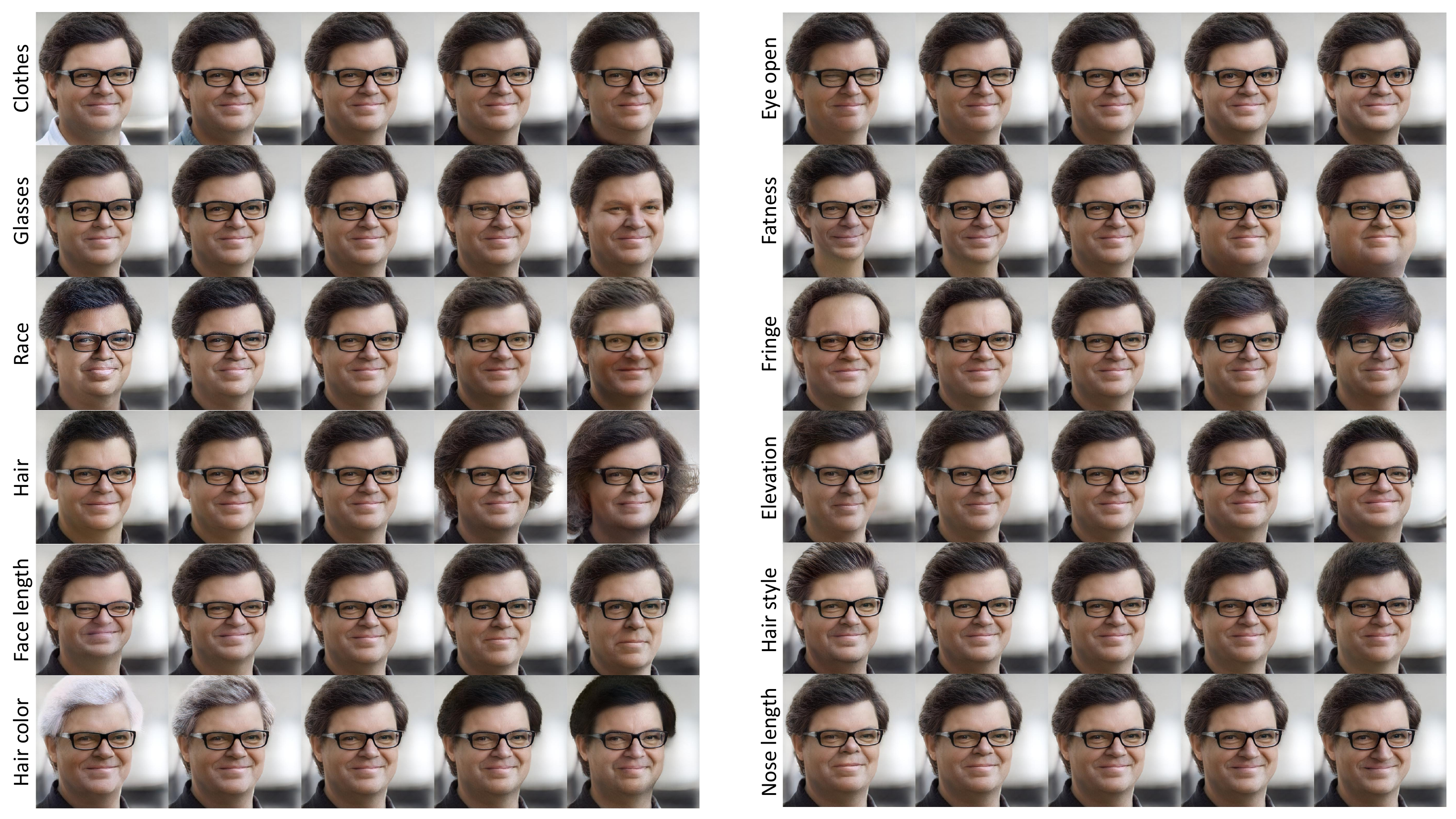}
\end{center}
    \caption{Real-image edit 2.}
\label{fig:lecun}
\end{figure}
\begin{figure}
\begin{center}
   \includegraphics[width=\linewidth]{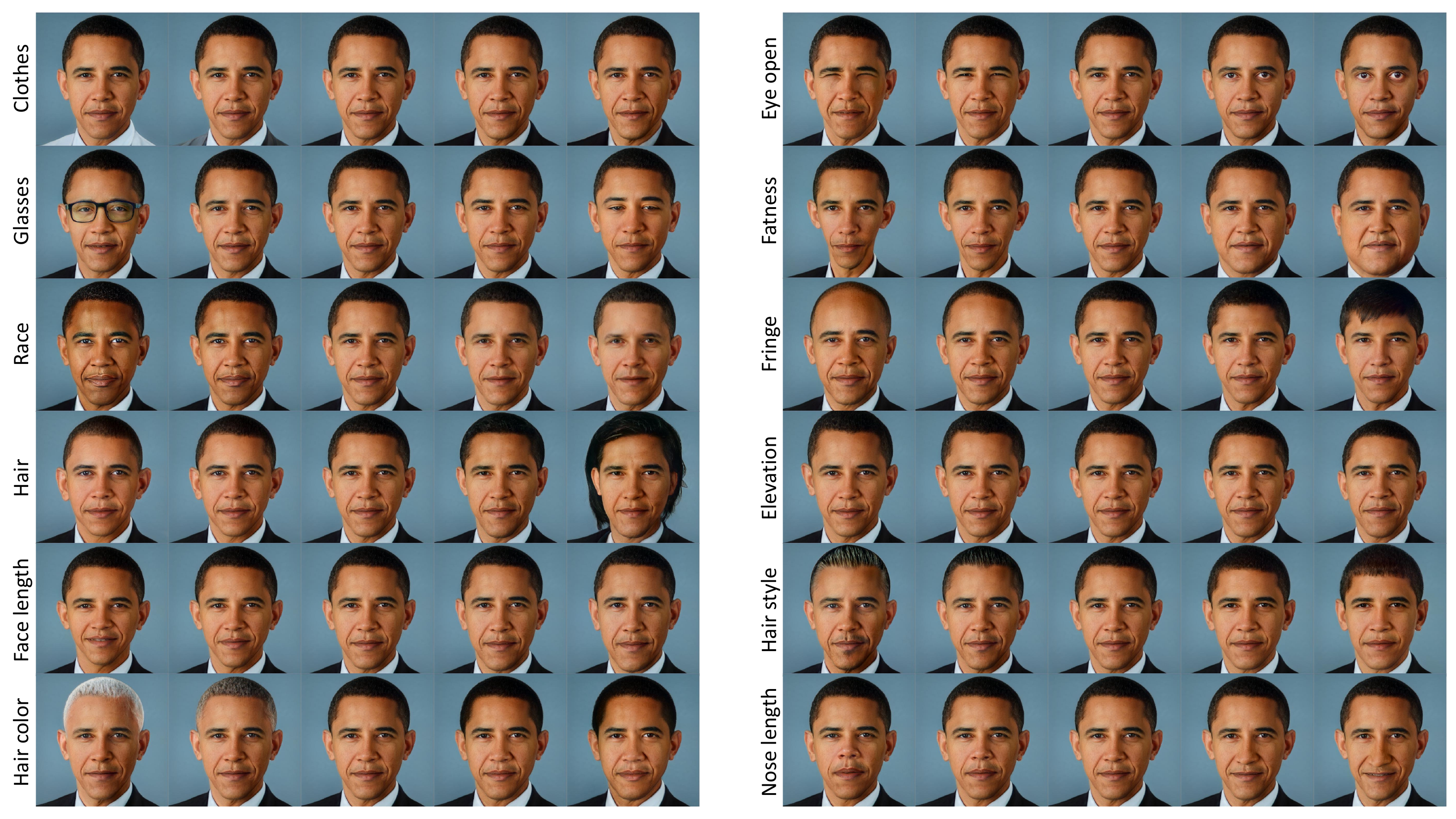}
\end{center}
    \caption{Real-image edit 3.}
\label{fig:obama}
\end{figure}
In Fig. \ref{fig:bill_gates}, Fig. \ref{fig:lecun}, and Fig. \ref{fig:obama}
we show the real-image on Gates, Lecun, and Obama.
Most of these semantic changes can be applied successfully and
consistently.
However, there still exist some problematic cases.
For example, the \emph{hair} semantic change applied to Obama
(Fig. \ref{fig:obama}) is not very satisfactory as there exists an obvious
identity shift while the hair length change is not very natural.
This phenomenon is due to the skewness of the latent space.
In our modeling, we learn the semantic directions to be
global directions shared by all positions in the latent space.
However, this perfect homogeneity assumption does not hold in reality,
leading to the Fig. \ref{fig:obama} case, where the \emph{hair} direction
affects not only the hair semantics but also other attributes.

\section{Other Datasets}
\label{ap:other_sets}
\begin{figure*}
\begin{center}
   \includegraphics[width=\linewidth]{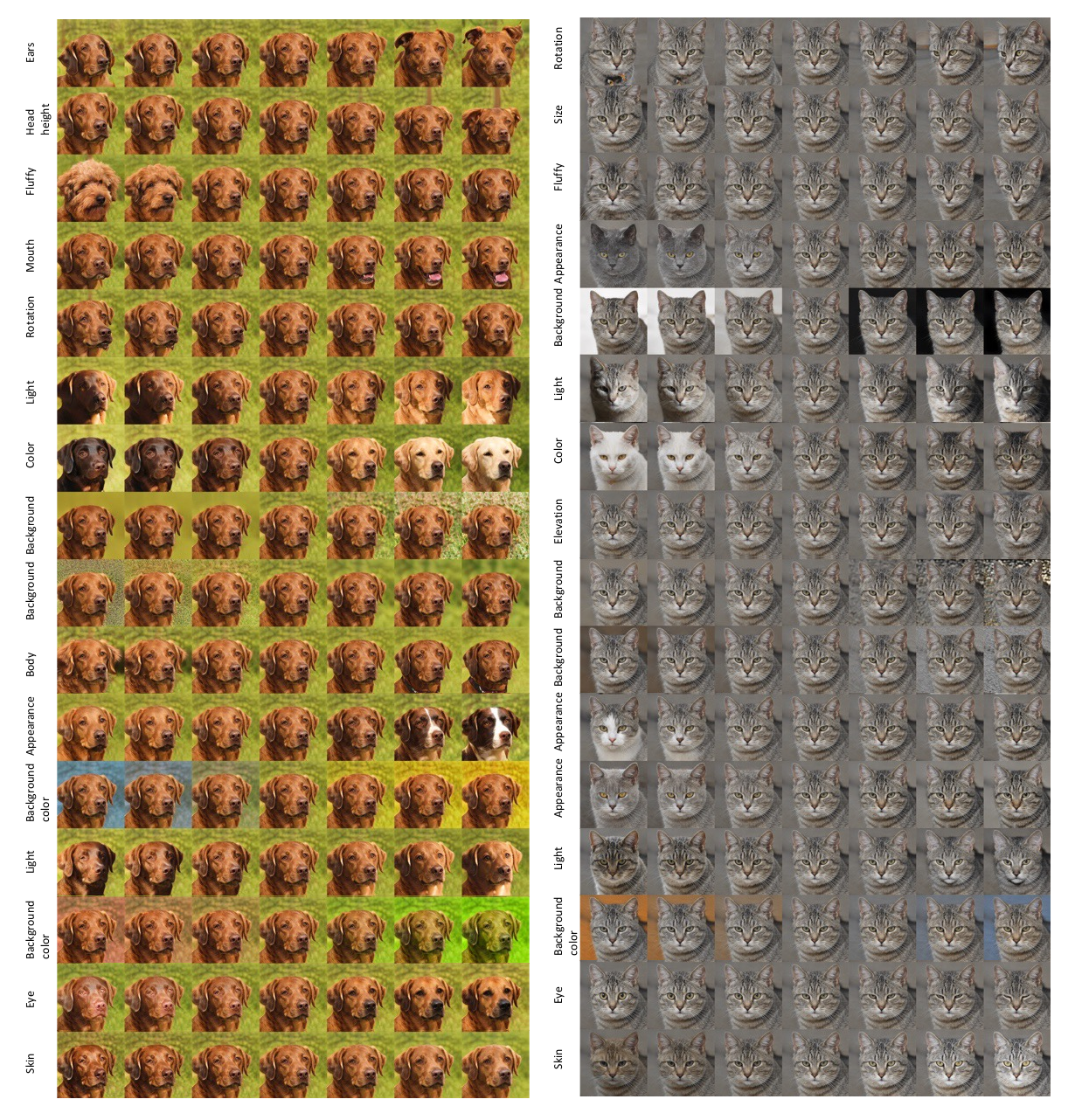}
\end{center}
    \caption{Discovered semantic changes on AFHQ-Dog and AFHQ-Cat datasets.}
\label{fig:dogs_cats}
\end{figure*}
\begin{figure*}
\begin{center}
   \includegraphics[width=\linewidth]{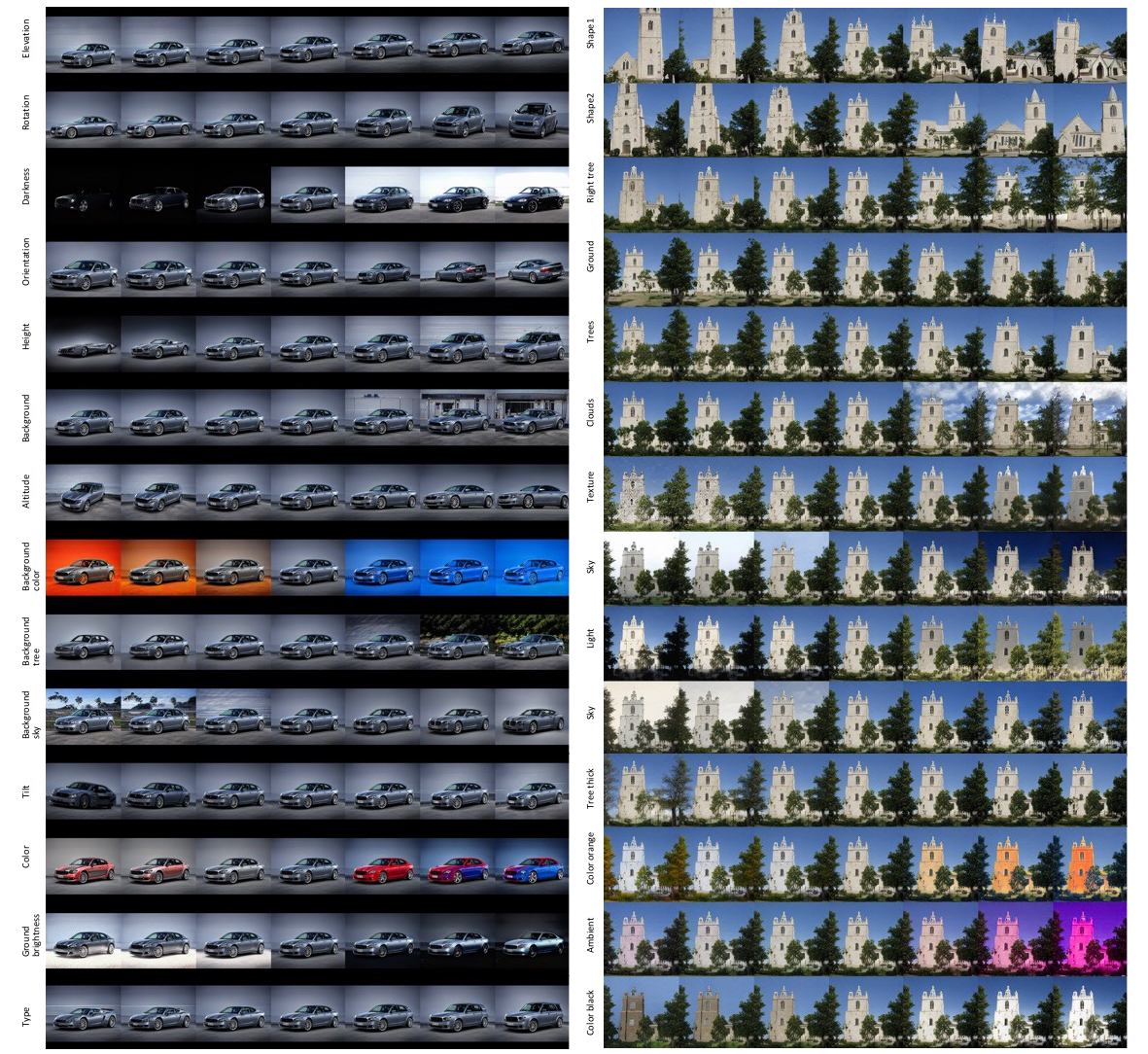}
\end{center}
    \caption{Discovered semantic changes on LSUN Car and Church datasets.}
\label{fig:cars_churches}
\end{figure*}
We show more discovered directions on AFHQ-Dog, Cat, LSUN Car, Church datasets
in Fig. \ref{fig:dogs_cats} and \ref{fig:cars_churches}.

\section{Qualitative Results on 3DShapes}
\label{ap:3dshapes_trav}
\begin{figure*}
\begin{center}
   \includegraphics[width=\linewidth]{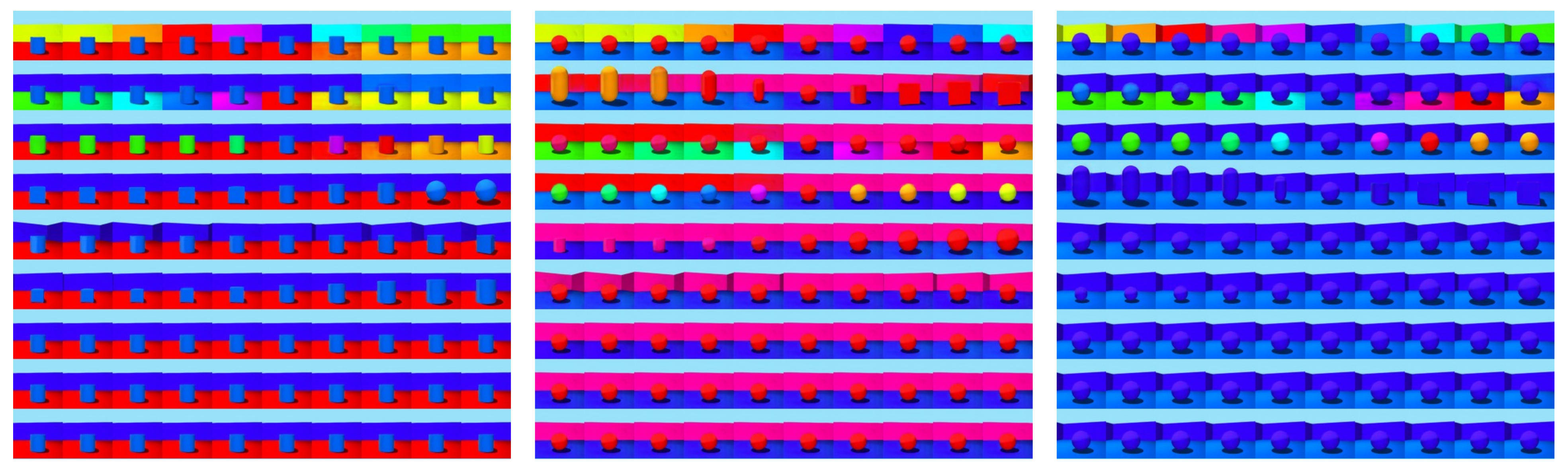}
\end{center}
    \caption{Discovered directions on 3DShapes dataset.}
\label{fig:shapes3d}
\end{figure*}

\begin{figure*}
\begin{center}
   \includegraphics[width=\linewidth]{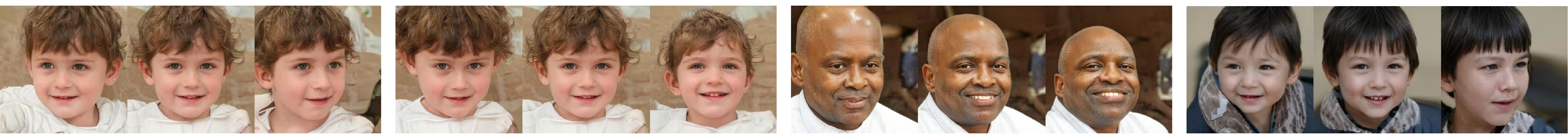}
\end{center}
    \caption{Failure cases. Some other attributes are correlated with rotation or elevation.}
\label{fig:failures}
\end{figure*}
The qualitative discovery results on 3DShapes are shown in
Fig. \ref{fig:shapes3d}.
We show qualitative traversal results on 3DShapes dataset
in Fig. \ref{fig:shapes3d}.

\section{Implementation Details}
\label{ap:implementations}
We use pretrained StyleGAN2 generators for all experiments in this paper
due to its outstanding performance in generation and disentanglement.
The used models are of 512x512 resolution on
FFHQ, AFHQ, LSUN Car datasets, 256x256 resolution on Church dataset, and 64x64
resolution on 3DShapes dataset.

We use the pretrained AlexNet backbone without FC classifier layers
as the deep feature extractor $E$ in all experiments.
The losses (consistency and orthogonality) are computed based on
multiple stages of the extracted feature maps by the feature extractor.

For the navigator module, we implement the two
branches (direction branch and attention
branch) to predict modifications in the $\mathcal{W}$ space
of StyleGANs.
The directions predicted by the direction branch are normalized
to be the same length, and the attention branch goes through a
softmax function and a Gaussian-kernel convolution
operation (kernel size is 3) as introduced in the main paper.

For the prototype variation pattern variant, we set the size of
variation patterns to be the same as the input images.
The learned patterns are fed through a $\mathtt{tanh}$ function
to keep the input range between [-1, 1]
before fed into the feature extractor.

We train our models for 600k images using Adam optimizer with betas 0 and 0.99.
The initial learning rates are set to 1.

For the disentangled representation model used in our proposed metrics,
we use the published pretrained FFHQ generator model in \cite{Xinqi_cvpr21}.
Because the generator is already disentangled with each latent dimension
controlling a single semantic change in the images, we train a
predictor $P$ to \emph{invert} this correspondence $\bsb{z} = P(\bsb{X})$.
The predictor $P$ embeds the image into a disentangled latent space
with each dimension representing a score for an attribute (learned by the
disentangled generator). The predictor is trained with images generated
from the disentangled generator by regression (fitting the latent code).

\section{Limitations}
\label{ap:falures}
Although our semantic discovery models have demonstrated better
performance than other existing models, there are still limitations
our models cannot perfectly handle.
There are two types of limitations that are most commonly observed.
The first type is the lack of 3D consistency. As shown in Fig. \ref{fig:failures},
sometimes the model will generate inconsistent rotations and elevations, where some
other attributes (like face shape) are also changed during rotation.
These failures are more obvious on infrequent samples like children faces.
We suspect this may be due to (1) the lack of 3D prior in the
StyleGAN architecture, and (2) the weak judgment capability on 3D consistency
in the feature extractor used in our models. We believe with the rise of
3D-based models (e.g. NeRF-based models) this 3D consistency issue
can be inherently solved soon.
Another type of limitation is the manual assignment of attribute names.
Currently the discovered attributes are assigned with a name in a post-hoc manner
where we manually inspect the variations controlled by a discovered direction and give 
each direction a name. It may be solvable by leveraging a vision-language model like
CLIP \cite{CLIPmodel} to accomplish this zero-shot name assignemnt task.

\end{document}